\documentclass[aoas,MSNbibl,nameyear,seceqn,rotating,dvips]{arximspdf}
\usepackage{algorithmic,algorithm}
\usepackage{dcolumn}
\usepackage{stfloats}
\usepackage{graphicx}

\doi{10.1214/11-AOAS514}
\volume{6}
\issue{2}
\pubyear{2012}
\firstpage{719}
\lastpage{752}

\makeatletter

\fnbelowfloat

\newcolumntype{d}[1]{D{.}{.}{#1}}

\newcommand{\argmin}{\mathop{\arg\min}}
\newcommand{\argmax}{\mathop{\arg\max}}

\newcommand{\bu}{{\mathbf u}}
\newcommand{\bv}{{\mathbf v}}
\newcommand{\bX}{{\mathbf X}}
\newcommand{\bY}{{\mathbf Y}}
\newcommand{\bB}{{\mathbf B}}
\newcommand{\bA}{{\mathbf A}}
\newcommand{\giG}{{g \in\mathcal{G}}} 
\newcommand{\be}{{\bolds\varepsilon}}
\newcommand{\bw}{{\mathbf w}}

\newtheorem{proposition}{Proposition}
\newtheorem{lemma}{Lemma}

\newproclaim{definition}{Definition}

\newtheorem{theorem}{Theorem}

\newproclaim{remark}{Remark}

\makeatother

\begin{document}
\begin{frontmatter}

\title{Smoothing proximal gradient method for general structured
sparse regression}
\runtitle{Smoothing proximal gradient method}

\begin{aug}
\author[A]{\fnms{Xi} \snm{Chen}\ead[label=e1]{xichen@cs.cmu.edu}},
\author[B]{\fnms{Qihang} \snm{Lin}\ead[label=e2]{qihangl@andrew.cmu.edu}},
\author[A]{\fnms{Seyoung} \snm{Kim}\ead[label=e3]{sssykim@cs.cmu.edu}},
\author[A]{\fnms{Jaime~G.}~\snm{Carbonell}\ead[label=e4]{jgc@cs.cmu.edu}}
\and
\author[A]{\fnms{Eric~P.}~\snm{Xing}\corref{}\thanksref{t1}\ead[label=e5]{epxing@cs.cmu.edu}}
\runauthor{X. Chen et al.}
\affiliation{Carnegie Mellon University}
\address[A]{X. Chen\\
S. Kim\\
J. G. Carbonell\\
E. P. Xing\\
School of Computer Science\\
Carnegie Mellon University\\
5000 Forbes Avenue\\
Pittsburgh, Pennsylvania 15213-3891\\
USA\\
\printead{e1}\\
\phantom{E-mail: }\printead*{e3}\\
\phantom{E-mail: }\printead*{e4}\\
\phantom{E-mail: }\printead*{e5}}
\address[B]{Q. Lin\\
Tepper School of Business\\
Carnegie Mellon University\\
5000 Forbes Avenue\\
Pittsburgh, Pennsylvania 15213-3891\\
USA\\
\printead{e2}} 
\end{aug}

\thankstext{t1}{Supported by Grants ONR
N000140910758, NSF DBI-0640543, NSF CCF-0523757, NIH 1R01GM087694,
AFOSR FA95501010247, NIH 1R01GM093156 and an Alfred P. Sloan Research
Fellowship awarded to EPX.}

\received{\smonth{3} \syear{2011}}
\revised{\smonth{7} \syear{2011}}

%
\begin{abstract}
We study the problem of estimating high-dimensional regression
models regularized by a structured sparsity-inducing penalty that
encodes prior structural information on either the input or output
variables. We consider two widely adopted types of penalties of this
kind as
motivating examples: (1) the general overlapping-group-lasso penalty,
generalized from the group-lasso penalty;
and (2) the graph-guided-fused-lasso penalty, generalized from the
fused-lasso penalty. For both types of penalties, due to their
nonseparability and nonsmoothness, developing an efficient optimization
method remains a~challenging problem. In this paper we propose a
general optimization approach, the \textit{smoothing proximal gradient}
(SPG) method, which can solve structured sparse regression problems
with any smooth convex loss under a wide spectrum of structured
sparsity-inducing penalties. Our approach combines a smoothing
technique with an effective proximal gradient method. It achieves a
convergence rate significantly faster than the standard first-order
methods, subgradient methods, and is much more scalable than the most
widely used interior-point methods. The efficiency and scalability of
our method are demonstrated on both simulation experiments and real
genetic data sets.
\end{abstract}

%
\begin{keyword}
\kwd{Sparse regression}
\kwd{structured sparsity}
\kwd{smoothing}
\kwd{proximal gradient}
\kwd{optimization}.
\end{keyword}

\end{frontmatter}

\section{Introduction}

The problem of high-dimensional sparse feature learning arises in many
areas in science and engineering. In a typical setting such as linear
regression, the input signal leading to a response (i.e., the output)
lies in a high-dimensional space, and one is interested in selecting a
small number of truly relevant variables in the input that influence
the output. A~popular approach to achieve this goal is to jointly
optimize the fitness loss function with a nonsmooth $\ell_1$-norm
penalty, for example, Lasso [\citet{Tibshirani96}] that shrinks the coefficients
of the irrelevant input variables to zero. However, this approach is
limited in that it is incapable of capturing any structural information
among the input variables. Recently, various extensions of the
$\ell_1$-norm lasso penalty have been introduced to take advantage of
the prior knowledge of the structures among inputs to encourage closely
related inputs to be selected jointly
[\citet{Bach09}, \citet{Tibshirani05}, \citet{Yuan06}].
Similar ideas have also been explored to
leverage the output structures in multivariate-response regression (or
multi-task regression), where one is interested in estimating multiple
related functional mappings from a common input space to multiple
outputs [Kim and Xing (\citeyear{graph0901,Seyoung09}),
\citet{Obozinski09}]. In this case, the
structure over the outputs is available as prior knowledge, and the
closely related outputs according to this structure are encouraged to
share a similar set of relevant inputs. These progresses
notwithstanding, the development of efficient optimization methods for
solving the estimation problems resultant from the \textit{structured
sparsity-inducing penalty functions} remains a challenge for reasons we
will discuss below. In this paper we address the problem of developing
efficient optimization methods that can handle a broad family of
structured sparsity-inducing penalties with complex structures.

When the structure to be imposed during shrinkage has a relatively
simple form, such
as nonoverlapping groups over variables (e.g., group lasso
[\citet{Yuan06}]) or a linear-ordering (a.k.a., chain) of variables
(e.g., fused lasso [\citet{Tibshirani05}]), efficient optimization
methods have been developed. For example, under group lasso, due to the
separability among groups, a
\textit{proximal operator}\setcounter{footnote}{1}\footnote{The proximal operator associated
with the penalty is defined as $\argmin_{{\bolds\beta}} \frac{1}{2}
\|{\bolds\beta}-\bv\|_2^2+P({\bolds\beta})$, where $\bv
$ is any given vector and
$P({\bolds\beta})$ is the nonsmooth penalty.} associated with
the penalty
can be computed in closed-form; thus, a number of composite
gradient methods [\citet{FISTA}, \citet{Liu09multi-taskfeature},
\citet{Nesterov07}]
that leverage the proximal operator as a key step (so-called
``proximal gradient method'') can be directly applied.
For fused lasso, although the penalty is not separable, a coordinate
descent algorithm was shown feasible by explicitly leveraging the
linear ordering of the inputs [\citet{Friedman07}].

Unfortunately, these algorithmic advancements have been outpaced by the
emergence of more complex structures one would like to impose during
shrinkage. For example, in order to handle a more general class of
structures such as a tree or a graph over variables, various regression
models that further extend the group lasso and fused lasso ideas have
been recently proposed. Specifically, rather than assuming the variable
groups to be nonoverlapping as in the standard group lasso, the
\textit{overlapping group lasso} [\citet{Bach09}] allows each input variable to
belong to multiple groups, thereby introducing overlaps among groups
and enabling incorporation of more complex prior knowledge on the
structure. Going beyond the standard fused lasso, the
\textit{graph-guided fused lasso} extends the original chain structure
over variables to a general graph over variables, where the fused-lasso
penalty is applied to each edge of the graph [\citet{graph09}]. Due to
the nonseparability of the penalty terms resultant from the overlapping
group or graph structures in these new models, the aforementioned fast
optimization methods originally tailored for the standard group lasso
or fused lasso cannot be readily applied here, due to, for example,
unavailability of a closed-form solution of the proximal operator. In
principle, generic convex optimization solvers such as the
interior-point methods (IPM) could always be used to solve either a
second-order cone programming (SOCP) or a quadratic programming (QP)
formulation of the aforementioned problems; but such approaches are
computationally prohibitive for problems of even a moderate size. Very
recently, a great deal of attention has been given to devise practical
solutions to the complex structured sparse regression problems
discussed above in statistics and the machine learning community, and
numerous methods have been proposed [\citet{Duchi09},
\citet{Bach10}, \citet{Jieping10}, \citet{SSparse10},
\citet{Ryan10}, \citet{Hua11a}]. All of these recent works
strived to provide clever solutions to various subclasses of the
structured sparsity-inducing penalties; but, as we survey in Section
\ref{secsummary}, they are still short of reaching a simple, unified
and general solution to a broad class of structured sparse regression
problems.

In this paper we propose a generic optimization approach, the
\textit{smoothing proximal gradient} (SPG) method, for dealing with a broad
family of sparsity-inducing penalties of complex structures. We use the
overlapping-group-lasso penalty and graph-guided-fused-lasso penalty
mentioned above as our motivating examples. Although these two types of
penalties are seemingly very different, we show that it is possible to
decouple the nonseparable terms in both penalties via the dual norm;
and reformulate them into a~common form to which the proposed method
can be applied. We call our approach a ``smoothing'' proximal
gradient method because instead of optimizing the original objective function
directly as in other proximal gradient methods, we introduce a
\textit{smooth} approximation to the structured sparsity-inducing
penalty using the technique from \citet{Nesterov05}. Then, we solve
the smoothed surrogate problem by a first-order proximal gradient
method known as the fast iterative shrinkage-thresholding algorithm
(FISTA) [\citet{FISTA}]. We show that although we solve a smoothed
problem, when the smoothness parameter is carefully chosen, SPG
achieves a convergence
rate of $O(\frac{1}{\varepsilon})$ for the original objective for any
desired accuracy $\varepsilon$. Below, we summarize the main advantages
of this approach:

\begin{longlist}[(a)]
\item[(a)] It is a first-order method, as it uses
only the gradient information. Thus, it is significantly more
scalable than IPM for SOCP or QP. Since it is gradient-based, it
allows warm restarts, and thereby potentiates solving the problem along
the entire regularization path [\citet{Friedman07}].

\item[(b)] It is applicable to a wide class of optimization problems with
a smooth convex loss and a nonsmooth nonseparable structured
sparsity-inducing penalty. Additionally, it is applicable to both uni-
and multi-task sparse structured regression, with structures on either
(or both) inputs/outputs.

\item[(c)] Theoretically, it enjoys a convergence rate of $O(\frac
{1}{\varepsilon})$,
which dominates that of the standard first-order method such as the subgradient
method whose rate is of $O(\frac{1}{\varepsilon^2})$.\vspace*{1pt}

\item[(d)] Finally, SPG is easy to implement with a few lines of MATLAB code.
\end{longlist}

The idea of constructing a smoothing approximation to a
difficult-to-optimize objective function has also been adopted in
another widely used optimization
framework known as majorization--minimization (MM) [\citet{MM}]. Using
the quadratic surrogate functions for the $\ell_2$-norm and
fused-lasso penalty as derived in
\citet{MMLasso} and \citet{MMFused}, one can also apply MM to solve the
structured sparse regression problems.
We will discuss in detail the connections between our methods and MM in
Section \ref{secsummary}.

The rest of this paper is organized as follows. In Section
\ref{secback} we present the formulation of overlapping group
lasso and graph-guided fused lasso. In Section \ref{secspg} we
present the SPG method along with complexity results. In Section
\ref{secsummary} we discuss the connections between our method and
MM, and comparisons with other related methods. In Section
\ref{secmulti} we extend our algorithm to multivariate-task
regression. In Section \ref{secexp} we present numerical results
on both simulated and real data sets, followed by conclusions in
Section \ref{secconclusion}. Throughout the paper, we will discuss
overlapping-group-lasso and graph-guided-fused-lasso penalties in
parallel to illustrate how the SPG can be used to solve the
corresponding optimization problems generically.

\section{Background: Linear regression regularized by structured
sparsity-\break inducing penalties}
\label{secback}

We begin with a basic outline of the high-dimensional linear regression
model, regularized by structured sparsity-inducing penalties. 

Consider a data set of $N$ feature/response (i.e., input/output) pairs,
$\{{\mathbf x}_n,y_n\}$, $n=1, \ldots, N$. Let $\bX\in\mathbb{R}^{N
\times J}$ denote the matrix of inputs of the~$N$
samples, where each sample lies in a $J$-dimensional space; and
${\mathbf y}\in\mathbb{R}^{N \times1}$ denote the vector of
univariate outputs of the $N$ sample. Under a linear
regression model, $ {\mathbf y}=\bX{\bolds\beta}+ \be, $ where
${\bolds\beta}$ represents the vector of
length $J$ for the regression coefficients, and $\be$ is the vector of
length $N$ for
noise distributed as $N(0, \sigma^2 I_{N \times N})$. The well-known
Lasso regression [\citet{Tibshirani96}] obtains a sparse estimate of\vadjust{\goodbreak}
the coefficients
by solving the following optimization problem:
%
\begin{equation}
\min_{{\bolds\beta}\in\mathbb{R}^J} g({\bolds\beta
})+\lambda\|{\bolds\beta}\|_1,
\end{equation}
where $g({\bolds\beta}) \equiv\frac{1}{2} \|{\mathbf y}-\bX
{\bolds\beta}\|_2^2$ is the
squared-error loss, $\|{\bolds\beta}\|_1 \equiv{\sum
_{j=1}^J}|\beta_j|$ is the
$\ell_1$-norm penalty that encourages the solutions to be sparse, and
$\lambda$ is the regularization parameter that controls the
sparsity level.

The standard lasso penalty does not assume any structure among the
input variables, which limits its applicability to complex
high-dimensional scenarios in many applied problems. More structured
constraints on the input variables such as groupness or pairwise
similarities can be introduced by employing a more sophisticated
sparsity-inducing penalty that induces joint sparsity patterns among
related inputs. We generically denote the structured sparsity-inducing
penalty by $\Omega({\bolds\beta})$ without assuming a specific
form, and define the problem of estimating a
structured sparsity pattern of the coefficients as follows:
%
\begin{equation}\label{eqobjglasso}
\min_{{\bolds\beta}\in\mathbb{R}^J} f({\bolds\beta})
\equiv g({\bolds\beta})
+\Omega({\bolds\beta})+\lambda\|{\bolds\beta}\|_1.
\end{equation}

In this paper we consider two types of $\Omega({\bolds\beta})$
that capture two different kinds of structural constraints over
variables, namely, the overlapping-group-lasso penalty based on the
$\ell_1/\ell_2$ mixed-norm, and the graph-guided-fused-lasso penalty
based on a total variation norm. As we discuss below, these two types
of penalties represent a broad family of structured sparsity-inducing
penalties recently introduced in the literature
[\citet{Bach09}, \citet{Seyoung09}, \citet{graph09}, \citet{Tibshirani05}, \citet{Yuan06},
\citet{BinYu09}]. It is
noteworthy that in problem (\ref{eqobjglasso}), in addition to the
structured-sparsity-inducing penalty $\Omega({\bolds\beta})$,
there is also an $\ell_1$-regularizer $\lambda\|{\bolds\beta}\|
_1$ that explicitly enforces sparsity on every individual feature. The
SPG optimization algorithm to be presented in this paper is applicable
regardless of the presence or absence of the $\lambda\|{\bolds
\beta}\|_1$ term.

%
\begin{longlist}[(2)]

\item[(1)] \textit{Overlapping-group-lasso penalty.}
Given prior knowledge of (possibly overlapping) grouping of variables
or features, if it is desirable to encourage coefficients of features
within the same group to be shrunk to zero jointly, then a composite
structured penalty of the following form can be used:
%
\begin{equation}
\label{eqpennormgroup}
\Omega({\bolds\beta}) \equiv\gamma\sum_{g\in\mathcal{G}}
w_g\|
{\bolds\beta}_g\|_2,
\end{equation}
where $\mathcal{G}=\{g_1, \ldots, g_{|\mathcal{G}|}\}$ denotes the
set of groups, which is a subset of the power set of $\{1, \ldots, J\}
$; ${\bolds\beta}_g\in\mathbb{R}^{|g|}$ is the subvector of
${\bolds\beta}$ for the
features in group $g$; $w_g$~is the predefined weight for group $g$;
and \mbox{$\|\cdot\|_2$} is the vector $\ell_2$-norm.\vadjust{\goodbreak} This $\ell_1/\ell_2$
mixed-norm penalty plays the role of jointly setting all of the
coefficients within each group
to zero or nonzero values. The widely used hierarchical tree-structured
penalty [\citet{Seyoung09}, \citet{iCAP08}] is a special case of (\ref{eqpennormgroup}),
of which the groups are defined as a~nested set under a tree hierarchy.
It is noteworthy that the $\ell_1/\ell_{\infty}$ mixed-norm penalty
can also achieve a similar grouping effect.
Indeed, our approach can also be applied to the $\ell_1/\ell_{\infty
}$ penalty, but for simplicity here we focus on only the $\ell_1/\ell
_2$ penalty and the comparison
between the $\ell_1/\ell_2$ and the $\ell_1/\ell_{\infty}$ is
beyond the scope of the paper.

Apparently, the penalty $\Omega({\bolds\beta}) \equiv\gamma
\sum_{g\in\mathcal{G}} w_g\| {\bolds\beta}_g\|_2$ alone
enforces only group-level sparsity but not\vspace*{1pt} sparsity within each group.
More precisely, if the estimated $\|\widehat{{\bolds\beta}}_g\|
_2 \neq0$, each $\widehat{\beta}_j$ for $j \in g$ will be nonzero.
By using an additional $\ell_1$-regularizer $\lambda\|{\bolds
\beta}\|_1$ together with $\Omega({\bolds\beta})$ as in (\ref
{eqobjglasso}), one cannot only select groups but also variables within
each group. The readers may refer to \citet{Friedman10b} for more details.


%
\item[(2)] \textit{Graph-guided-fused-lasso penalty.}
Alternatively, prior knowledge about the structural constraints over
features can be in the form of their pairwise relatedness described by
a graph $G \equiv(V, E)$, where $V=\{1,\ldots, J\}$ denotes the
variables or features of interest, and $E$ denotes the set of edges
among~$V$.
Additionally, we let $r_{ml} \in\mathbb{R}$ denote the weight of the
edge $e=(m,l) \in E$, corresponding to correlation or other proper
similarity measures between features $m$ and $l$.
If it is desirable to encourage coefficients of related features to
share similar magnitude, then the graph-guided-fused-lasso
penalty [\citet{graph09}] of the following form can be used:
%
\begin{equation}\label{eqpennormgraph}
\Omega({\bolds\beta})= \gamma\sum_{e=(m,l) \in E, m<l } \tau(r_{ml})
|\beta_m-\operatorname{sign}(r_{ml})\beta_l|,
\end{equation}
where $\tau(r_{ml})$ represent a general weight function that enforces
a fusion effect over coefficients $\beta_{m}$ and $\beta_{l}$ of
relevant features.
In this paper we consider $\tau(r)=|r|$, but any monotonically
increasing function of the absolute values of correlations can be used.

The $\operatorname{sign}(r_{ml})$ in (\ref{eqpennormgraph}) ensures
that two positively correlated inputs would tend to influence the
output in the same direction,
whereas two negatively correlated inputs impose opposite effect.
Since the fusion effect is calibrated by the edge weight, the
graph-guided-fused-lasso penalty in (\ref{eqpennormgraph}) encourages highly
inter-correlated inputs corresponding to a densely connected subnetwork
in $G$ to be jointly selected as relevant.

It is noteworthy that when $r_{ml}=1$ for all $e=(m,l) \in E$, and $G$
is simply a chain over nodes, we have
%
\begin{equation}\label{eqpennormgraphsimple}
\Omega({\bolds\beta})=\gamma\sum_{j=1}^{J-1}
|\beta_{j+1}-\beta_j|,
\end{equation}
which is identical to the standard fused lasso penalty
[\citet{Tibshirani05}].\vadjust{\goodbreak}
\end{longlist}

\section{Smoothing proximal gradient}
\label{secspg}
%

Although (\ref{eqobjglasso}) defines a convex program, of which
a globally optimal solution to ${\bolds\beta}$ is attainable,
the main
difficulty in solving (\ref{eqobjglasso}) arises from the
nonseparability of elements of ${\bolds\beta}$ in the nonsmooth penalty
function $\Omega({\bolds\beta})$. As we show in the next
subsection, although
the overlapping-group-lasso and graph-guided-fused-lasso penalties
are seemingly very different, we can reformulate the two types of
penalties as a common matrix algebraic form, to which a generic
Nesterov smoothing technique can be applied. The key in our approach
is to decouple the nonseparable structured sparsity-inducing
penalties into a simple linear transformation of ${\bolds\beta}$
via the dual
norm. Based on that, we introduce a smooth approximation to
$\Omega({\bolds\beta})$ using the technique from \citet
{Nesterov05} such that
its gradient with respect to ${\bolds\beta}$ can be easily calculated.

\subsection{Reformulation of structured sparsity-inducing
penalty} \label{secreform}
%

In this section we show that utilizing the dual norm, the
nonseparable structured sparsity-inducing penalty in both
(\ref{eqpennormgroup}) and (\ref{eqpennormgraph}) can be
decoupled; and reformulated into a common form as a maximization
problem over the auxiliary variables.

\begin{longlist}[(1)]
\item[(1)] \textit{Reformulating overlapping-group-lasso penalty.}
%
%
Since the dual norm of an $\ell_2$-norm is also $\ell_2$-norm, we
can write $\|{\bolds\beta}_g\|_2$ as $\|{\bolds\beta}_g\|
_2 = \max_{\|{\bolds\alpha}_g\|_2 \leq1}
{\bolds\alpha}_g^T{\bolds\beta}_g$, where ${\bolds
\alpha}_g \in\mathbb{R}^{|g|}$ is a vector of
auxiliary variables associated with ${\bolds\beta}_g$. Let
${\bolds\alpha}=[{\bolds\alpha}_{g_1}^T, \ldots,
{\bolds\alpha}_{g_{|\mathcal{G}|}}^T]^T$.
Then, ${\bolds\alpha}$ is a vector of length $\sum_{g \in G}
|g|$ with domain
$\mathcal{Q} \equiv\{{\bolds\alpha} | \|{\bolds\alpha
}_g\|_2 \leq1 ,  \forall g \in
\mathcal{G} \}$, where $\mathcal{Q}$ is the Cartesian product of
unit balls in Euclidean space and, therefore, a closed and convex set. We
can rewrite the overlapping-group-lasso penalty in
(\ref{eqpennormgroup}) as
%
\begin{equation}
\label{eqsaddlepengroup}\quad
\Omega({\bolds\beta})= \gamma\sum_{g \in\mathcal{G}} w_g
\max_{\|{\bolds\alpha}_g\|_2 \leq1}
{\bolds\alpha}_g^T{\bolds\beta}_g =\max_{{\bolds
\alpha}\in\mathcal{Q}} \sum_{g \in\mathcal{G}} \gamma w_g
{\bolds\alpha}_g^T{\bolds\beta}_g =\max_{{\bolds
\alpha}\in\mathcal{Q}} {\bolds\alpha}^T C {\bolds\beta},
\end{equation}
where $C \in\mathbb{R}^{{\sum_{g \in\mathcal{G}}|g|} \times J}$ is
a matrix defined as follows. The rows of $C$ are indexed by all
pairs of $(i,g) \in\{(i,g)| i\in g, i\in\{1,\ldots, J\}, g \in
\mathcal{G}\}$, the
columns are indexed by $j \in\{1, \ldots, J\}$, and each element of
$C$ is given as
%
\begin{equation}
\label{eqmatrixC} C_{(i,g),j}=\cases{
\gamma w_g, &\quad if $i=j$,\cr
0, &\quad otherwise.}
\end{equation}
%
%

Note that $C$ is a highly sparse matrix with only a single nonzero
element in each row and $\sum_{g \in\mathcal{G}}|g|$ nonzero
elements in the entire matrix, and, hence, can be stored with only a small
amount of memory during the optimization procedure.


\item[(2)] \textit{Reformulating graph-guided-fused-lasso penalty.}
First, we rewrite the graph-guided-fused-lasso penalty in
(\ref{eqpennormgraph}) as follows:
\[
\gamma\sum_{e=(m,l)\in E, m<l }\tau(r_{ml})
|\beta_m-\operatorname{sign}(r_{ml})\beta_l| \equiv\|C {\bolds\beta
}\|_1,
\]
where $C \in\mathbb{R}^{|E| \times J}$ is the edge-vertex incident
matrix:
%
\begin{equation}
\label{eqmatrixH} C_{e=(m,l), j}=\cases{
\gamma\cdot\tau(r_{ml}), &\quad if $j=m$,\cr
-\gamma\cdot\operatorname{sign}(r_{ml})\tau(r_{ml}), &\quad if $j=l$,\cr
0, &\quad otherwise.}
\end{equation}
Again, we note that $C$ is a highly sparse matrix with $2\cdot|E|$
nonzero elements. Since the dual norm of the $\ell_\infty$-norm is
the $\ell_1$-norm,
we can further rewrite the graph-guided-fused-lasso penalty as
%
\begin{equation}
\label{eqdualnorm}
\|C {\bolds\beta}\|_1 \equiv\max_{\|{\bolds\alpha}\|
_{\infty}\leq1} {\bolds\alpha}^T C {\bolds\beta},
\end{equation}
where ${\bolds\alpha}\in\mathcal{Q}=\{{\bolds\alpha}| \|
{\bolds\alpha}\|_{\infty} \leq1, {\bolds\alpha}\in
\mathbb{R}^{|E|}\}$ is a vector of auxiliary variables associated with
$\|C {\bolds\beta}
\|_1$, and $\|\cdot\|_\infty$ is the $\ell_\infty$-norm defined as
the maximum absolute value of all entries in the vector.
%
\begin{remark}
As a generalization of the graph-guided-fused-lasso penalty, the proposed
optimization method can be applied to the $\ell_1$-norm of any
linear mapping of ${\bolds\beta}$ [i.e., $\Omega({\bolds
\beta})=\|C{\bolds\beta}\|_1$ for any
given $C$].
\end{remark}
\end{longlist}
%

\subsection{Smooth approximation to structured sparsity-inducing penalty}

The common formulation of $\Omega({\bolds\beta})$ given above
[i.e., $\Omega({\bolds\beta})=
\max_{{\bolds\alpha}\in\mathcal{Q}} {\bolds\alpha}^T C
{\bolds\beta}$] is still a nonsmooth
function of ${\bolds\beta}$, and this makes the optimization
challenging. To
tackle this problem, using the technique from \citet{Nesterov05}, we
construct a smooth approximation to $\Omega({\bolds\beta})$ as follows:
%
\begin{equation}
\label{eqfmu}
f_\mu({\bolds\beta})=\max_{{\bolds\alpha}\in\mathcal{Q}}
\bigl({\bolds\alpha}^T C {\bolds\beta}- \mu\,
d({\bolds\alpha})\bigr),
\end{equation}
where $\mu$ is a positive smoothness parameter and $d({\bolds
\alpha})$ is a smoothing function
defined as $\frac{1}{2} \|{\bolds\alpha}\|_2^2$. The original
penalty term can
be viewed as $f_\mu({\bolds\beta})$ with $\mu=0$; and one can
verify that
$f_\mu({\bolds\beta})$ is a lower bound of $f_0({\bolds
\beta})$. In order to bound the
gap between $f_\mu({\bolds\beta})$ and $f_0({\bolds\beta
})$, let $D= \max_{{\bolds\alpha}\in
\mathcal{Q}} d({\bolds\alpha})$. In our problems, $D=|\mathcal
{G}|/2$ for the
overlapping-group-lasso penalty and $D=|E|/2$ for the graph-guided-fused-lasso
penalty. Then, it is easy to verify that the maximum gap
between $f_\mu({\bolds\beta})$ and $f_0({\bolds\beta})$
is $\mu D$:
\[
f_0 ({\bolds\beta}) - \mu D
\leq f_{\mu}({\bolds\beta}) \leq f_0 ({\bolds\beta}).
\]
From Theorem \ref{thmkey} as presented
below, we know that $f_\mu({\bolds\beta})$ is a smooth function
for any
$\mu>0$. Therefore, $f_\mu({\bolds\beta})$ can be viewed as a
\textit{smooth
approximation} to $f_0({\bolds\beta})$ with a maximum gap of
$\mu D$; and
the $\mu$ controls the gap between $f_\mu({\bolds\beta})$ and
$f_0({\bolds\beta})$.
Given a desired accuracy $\varepsilon$, the convergence result in
Section \ref{subseccomplexity} suggests $\mu=\frac{\varepsilon}{2D}$
to achieve the best convergence rate.

Now we present the key theorem [\citet{Nesterov05}] to show that
$f_\mu({\bolds\beta})$ is smooth in ${\bolds\beta}$ with
a simple form of the gradient.
%
\begin{theorem}\label{thmkey}
For any
$\mu>0$, $f_\mu({\bolds\beta})$ is a convex and
continuously-differen\-tiable
function in ${\bolds\beta}$, and the gradient of $f_\mu
({\bolds\beta})$ takes the
following form:
%
\begin{equation}
\label{eqgrad}
\nabla f_\mu({\bolds\beta})= C^T {\bolds\alpha}^{\ast},
\end{equation}
where ${\bolds\alpha}^{\ast}$ is the optimal solution to (\ref{eqfmu}).
Moreover, the gradient $\nabla f_\mu({\bolds\beta})$ is
Lipschitz continuous
with the Lipschitz constant $L_\mu=\frac{1}{\mu} \|C\|^2$,
where\vspace*{1pt}
$\|C\|$ is the matrix spectral norm of $C$ defined as $\|C\| \equiv
{\max_{\|\bv\|_2 \leq1}} \|C \bv\|_2$.
\end{theorem}

By viewing\vspace*{1pt} $f_\mu({\bolds\beta})$ as the \textit{Fenchel
conjugate} of $d(\cdot)$ at $\frac{C{\bolds\beta}}{\mu}$, the
smoothness can be obtained by applying Theorem 26.3 in \citet{Rock96}.
The gradient in (\ref{eqgrad}) can be derived from the Danskin's
theorem [\citet{Ber99}] and the Lipschitz constant is shown in \citet
{Nesterov05}. The details of the proof are given in the
\hyperref[app]{Appendix}.

\begin{figure}[b]
\begin{tabular}{@{}cc@{}}

\includegraphics{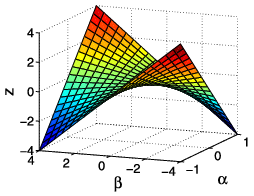}
 & \includegraphics{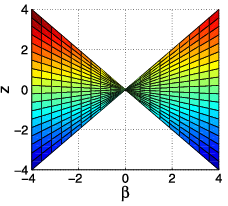}\\
(a) & (b) \\[4pt]

\includegraphics{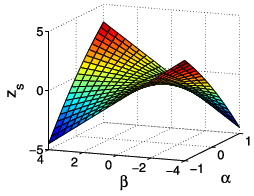}
 & \includegraphics{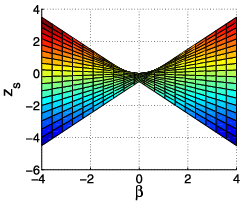} \\
(c) & (d)
\end{tabular}
\caption{A geometric illustration of the smoothness of $f_\mu(\beta)$.
\textup{(a)} The 3-D plot of $z(\alpha, \beta)$, \textup{(b)} the
projection of \textup{(a)} onto the $\beta$-$z$ space, \textup{(c)} the
3-D plot of $z_s(\alpha,\beta)$ and \textup{(d)} the projection of
\textup{(c)} onto the $\beta$-$z$ space.} \label{figsynvisual}
\end{figure}

\textit{Geometric illustration of Theorem} \ref{thmkey}. To provide
insights on why $f_{\mu}({\bolds\beta})$ is a smooth function
as Theorem \ref{thmkey}
suggests, in Figure \ref{figsynvisual} we show a geometric
illustration for the case of a one-dimensional parameter (i.e., $\beta
\in\mathbb{R}$) with $\mu$ and $C$ set to 1. First, we show
geometrically that $f_0(\beta)=\max_{\alpha\in[-1,1]}
z(\alpha,\beta)$ with $z(\alpha,\beta) \equiv\alpha\beta$ is a
nonsmooth function. The three-dimensional plot for $z(\alpha,\beta)
$ with $\alpha$ restricted to $[-1,1]$ is shown in Figure
\ref{figsynvisual}(a). We project the surface in Figure
\ref{figsynvisual}(a) onto the $\beta-z$ space as shown in Figure
\ref{figsynvisual}(b). For each~$\beta$, the value of $f_0(\beta)$
is the highest point along the $z$-axis since we maximize over
$\alpha$ in $[-1,1]$. We can see that $f_0(\beta)$ is composed of
two segments with a sharp point at $\beta=0$ and hence is
\textit{nonsmooth}. Now,\vspace*{1pt} we introduce $d(\alpha)=\frac{1}{2}\alpha^2$,
let $z_s(\alpha,\beta) \equiv\alpha\beta-\frac{1}{2} \alpha^2$ and
$f_\mu(\beta)=\max_{\alpha\in[-1,1]} z_s(\alpha,\beta)$. The
three-dimensional plot for $z_s(\alpha,\beta)$ with $\alpha$
restricted to $[-1,1]$ is shown in Figure \ref{figsynvisual}(c).
Similarly, we project the surface in Figure \ref{figsynvisual}(c)
onto the $\beta-z_s$ space as shown in Figure
\ref{figsynvisual}(d). For fixed $\beta$, the value of
$f_{\mu}(\beta)$ is the highest point along the $z$-axis. In Figure
\ref{figsynvisual}(d), we can see that the sharp point at
$\beta=0$ is removed and $f_{\mu}(\beta)$ becomes \textit{smooth}.

To compute the $\nabla f_\mu({\bolds\beta})$ and $L_\mu$, we
need to know
${\bolds\alpha}^{\ast}$ and $\|C\|$. We present the closed-form
equations for
${\bolds\alpha}^{\ast}$ and $\|C\|$ for the
overlapping-group-lasso penalty and
graph-guided-fused-lasso penalty in the following propositions. The
proof is presented in the \hyperref[app]{Appendix}.

\begin{longlist}[(1)]
\item[(1)] \textit{${\bolds\alpha}^{\ast}$ under
overlapping-group-lasso penalty.}
%
\begin{proposition}\label{propgroup}
Let ${\bolds\alpha}^{\ast}$, which is composed of $\{
{\bolds\alpha}^{\ast}_g\}_{g \in
\mathcal{G}}$, be the optimal solution to (\ref{eqfmu}) for
the overlapping-group-lasso penalty in (\ref{eqpennormgroup}). For
any $\giG$,
\[
{\bolds\alpha}^{\ast}_g= S\biggl(\frac{\gamma w_g {\bolds\beta
}_g}{\mu}\biggr),
\]
where
$S$ is the projection operator which projects any vector $\bu$ to
the $\ell_2$ ball:
\[
S(\bu) =
\cases{\dfrac{\bu}{\|\bu\|_2}, &\quad
$\|\bu\|_2 > 1$,\vspace*{2pt}\cr
\bu, &\quad $\|\bu\|_2 \leq1$.}
\]

In addition, we have $ \|C\|= \gamma\max_{j \in\{1, \ldots, J\}}
\sqrt{\sum_{\giG\ \mathrm{s.t.}\  j\in g} (w_g)^2}$.
\end{proposition}

\item[(2)] \textit{${\bolds\alpha}^{\ast}$ under
graph-guided-fused-lasso penalty.}
%
\begin{proposition}\label{propnormgraph}
Let ${\bolds\alpha}^{\ast}$ be the optimal solution of (\ref
{eqfmu}) for the graph-guided-fused-lasso
penalty in (\ref{eqpennormgraph}). Then, we have
\[
{\bolds\alpha}^{\ast}= S\biggl(\frac{ C {\bolds\beta}}{\mu}\biggr),
\]
where $S$ is the projection operator defined as follows:
\[
S(x)= \cases{
x, &\quad if $-1\leq x\leq1$,\cr
1, &\quad if $x>1$,\cr
-1, &\quad if $x<-1$.}
\]
For any vector ${\bolds\alpha}$, $S({\bolds\alpha})$ is
defined as applying $S$ on each
and every entry of~${\bolds\alpha}$.\vadjust{\goodbreak}

$\|C\|$ is upper-bounded by $\sqrt{2\gamma^2\max_{j \in V}d_j}$,
where
%
\begin{equation}
\label{eqd}
d_j=\sum_{e\in E \ \mathrm{s.t.} \  e \ \mathrm{incident}\  \mathrm{on} \
j}(\tau(r_e))^2
\end{equation}
for $j\in V$ in graph $G$, and this bound is tight.
Note that when $\tau(r_e)=1$ for all $e \in E$, $d_j$ is simply the
degree of the node $j$.
\end{proposition}
\end{longlist}

\subsection{Smoothing proximal gradient descent}

Given the smooth approximation to the nonsmooth structured
sparsity-inducing penalties, now, we apply the fast iterative
shrinkage-thresholding algorithm (FISTA) [\citet{FISTA}, \citet{Paul08}] to
solve a generically reformulated optimization problem, using the
gradient information from Theorem \ref{thmkey}. We substitute the
penalty term $\Omega({\bolds\beta})$ in (\ref{eqobjglasso})
with its smooth
approximation $f_\mu({\bolds\beta})$ to obtain the following
optimization
problem:
%
\begin{equation}
\min_{{\bolds\beta}} \widetilde{f}({\bolds\beta})
\equiv g({\bolds\beta})+ f_\mu({\bolds\beta}) +\lambda
\|{\bolds\beta}\|_1.
\end{equation}
Let
%
\begin{equation}\label{eqh}
h({\bolds\beta})=g({\bolds\beta})+ f_\mu({\bolds
\beta}) = \tfrac{1}{2}
\|{\mathbf y}-\bX{\bolds\beta}\|_2^2+f_{\mu}({\bolds
\beta})
\end{equation}
be the smooth part of $\widetilde{f}({\bolds\beta})$. According
to Theorem
\ref{thmkey}, the gradient of $h({\bolds\beta})$ is given as
%
\begin{equation}
\label{eqfgrad}
\nabla h ({\bolds\beta})= \bX^T(\bX{\bolds\beta}-
{\mathbf y}) + C^T {\bolds\alpha}^{\ast}.
\end{equation}
Moreover, $\nabla h({\bolds\beta})$ is Lipschitz-continuous with
the Lipschitz
constant,
%
\begin{equation}
\label{eqL}
L=\lambda_{\max} (\bX^T\bX)+ L_{\mu} = \lambda_{\max} (\bX^T\bX
) +
\frac{\|C\|^2}{\mu},
\end{equation}
where $\lambda_{\max} (\bX^T\bX)$ is the largest eigenvalue of
$(\bX^T\bX)$.\vspace*{1pt}

Since\vspace*{1pt} $\widetilde{f}({\bolds\beta})$ only involves a very simple
nonsmooth part (i.e., the $\ell_1$-norm penalty), we can\vspace*{1pt} adopt FISTA
[\citet{FISTA}, \citet{Paul08}] to minimize $\widetilde{f}({\bolds\beta})$ as
shown in Algorithm \ref{algogdglasso}. Algorithm \ref{algogdglasso}
alternates between the sequences $\{w^t\}$ and $\{{\bolds\beta}^t\}$
and $\theta_t$ can be viewed as a special ``step-size,'' which
determines the relationship between $\{w^t\}$ and $\{{\bolds\beta}^t\}$
as in Step 4 of Algorithm~\ref{algogdglasso}. As shown in
\citet{FISTA}, such a way of setting $\theta_t$ leads to Lemma
\ref{lemsmooth} in the \hyperref[app]{Appendix}, which further
guarantees the convergence result in Theorem \ref{thmcomplexity}.

\begin{algorithm}
\begin{flushleft}
\caption{Smoothing proximal gradient descent (SPG) for structured
sparse regression}\label{algogdglasso}
\textbf{Input}: $\bX$, ${\mathbf y}$, $C$, ${\bolds\beta}^0$,
Lipschitz constant $L$, desired accuracy $\varepsilon$.

\textbf{Initialization}: set $\mu=\frac{\varepsilon}{2D}$ where
$D=\max_{{\bolds\alpha}\in\mathcal{Q}} \frac{1}{2}\|
{\bolds\alpha}\|_2^2$
($D=|\mathcal{G}|/2$ for the overlapping-group-lasso\vspace*{1pt} penalty and
$D=|E|/2$ for the graph-guided-fused-lasso penalty), $\theta_0=1$,
$\bw^0={\bolds\beta}^0$.

\textbf{Iterate}: For $t=0,1,2,\ldots,$ until convergence of
${\bolds\beta}^t$:
\begin{longlist}[1.]
\item[1.] Compute $\nabla h(\bw^t)$ according to
(\ref{eqfgrad}).
\item[2.] Solve the proximal operator associated with the $\ell_1$-norm:
%
\begin{eqnarray}
\label{eqgradupdate}
{\bolds\beta}^{t+1} &=& \argmin
_{{\bolds\beta}} Q_L({\bolds\beta}, \bw^t) \nonumber\\[-8pt]\\[-8pt]
&\equiv& h(\bw
^t)+ \langle
{\bolds\beta}-\bw^t, \nabla h(\bw^t) \rangle+ \lambda\|
{\bolds\beta}\|_1
+\frac{L}{2}\|{\bolds\beta}-\bw^t\|_2^2.\nonumber
\end{eqnarray}
\item[3.] Set $\theta_{t+1}=\frac{2}{t+3}$.
\item[4.] Set $ \bw^{t+1} = {\bolds\beta}^{t+1}+\frac{1-\theta
_t}{\theta_t}\theta_{t+1}({\bolds\beta}^{t+1}-{\bolds
\beta}^t)$.
\end{longlist}
\textbf{Output}: $\widehat{{\bolds\beta}} ={\bolds\beta}^{t+1}$.
\end{flushleft}
\end{algorithm}

Rewriting $Q_L({\bolds\beta}, \bw^t)$ in (\ref{eqgradupdate}),
\[
Q_L({\bolds\beta}, \bw^t)= \frac{1}{2}\biggl\|{\bolds\beta
}-\biggl(\bw^t-\frac{1}{L}\nabla h(\bw^t)\biggr)\biggr\|_2^2
+ \frac{\lambda}{L} \|
{\bolds\beta}\|_1.
\]
Letting
$\bv=(\bw^t-\frac{1}{L}\nabla h(\bw^t))$, the closed-form solution
for ${\bolds\beta}^{t+1}$ can be obtained by soft-thresholding
[\citet{Friedman07}] as presented in the next proposition.
%
\begin{proposition}
\label{propsoftthreshold} The closed-form solution of
\[
\min_{{\bolds\beta}} \frac{1}{2}\|{\bolds\beta}-\bv\|
_2^2+\frac{\lambda}{L} \|{\bolds\beta}\|_1
\]
can be obtained by the soft-thresholding operation:
%
\begin{equation}
\label{eqsoftthreshold} \beta_j= \operatorname{sign}(v_j) \max\biggl(0,
|v_j|-\frac{\lambda}{L}\biggr), \qquad j=1, \ldots, J.
\end{equation}
\end{proposition}


An important advantage of using the proximal operator associated with
the $\ell_1$-norm $Q_L({\bolds\beta}, \bw^t)$ is that it can
provide us with sparse solutions, where the coefficients for irrelevant
inputs are set \textit{exactly} to zeros, due to the soft-thresholding
operation in (\ref{eqsoftthreshold}). 
When the term $\lambda\|{\bolds\beta}\|_1$ is not included in
the objective, for overlapping group lasso, we can only obtain the
group level sparsity but not the individual feature level sparsity
inside each group. However, as for optimization, Algorithm \ref
{algogdglasso} still applies in the same way. The only difference is
that Step 2 of Algorithm \ref{algogdglasso} becomes ${\bolds
\beta}^{t+1}=\argmin_{{\bolds\beta}} h(\bw^t)+ \langle
{\bolds\beta}-\bw^t, \nabla h(\bw^t) \rangle+\frac{L}{2}\|
{\bolds\beta}-\bw^t\|_2^2= \bw^t-\frac{1}{L}\nabla h(\bw
^t)$. Since there is no soft-thresholding step, the obtained solution
$\widehat{{\bolds\beta}}$ has no exact zeros. We then need to
set a threshold (e.g., $10^{-5}$) and select the relevant groups which
contain the variables with the parameter above this
threshold.\looseness=1


\subsection{Issues on the computation of the Lipschitz constant}
When $J$ is large, the computation  of $\lambda_{\max} (\bX^T\bX)$
and hence the Lipschitz constant $L$ could be very expensive. To
further accelerate Algorithm \ref{algogdglasso}, a line search
backtracking step could be used to dynamically assign a constant
$L_t$ for the proximal operator in each iteration [\citet{FISTA}]. More
specifically, given any positive constant~$R$, let
\[
Q_R({\bolds\beta},
\bw^t)= h(\bw^t)+ \langle{\bolds\beta}-\bw^t, \nabla h(\bw
^t) \rangle+
\lambda\|{\bolds\beta}\|_1 + \frac{R}{2} \|{\bolds\beta
}-\bw^t\|_2^2
\]
and
\[
{\bolds\beta}^{t+1} \equiv{\bolds\beta}_R(\bw^t)=
\argmin_{{\bolds\beta}} Q_R({\bolds\beta}, \bw^t).
\]
The key to guarantee the convergence rate of Algorithm
\ref{algogdglasso} is to ensure that the following inequality holds
for each iteration:
%
\begin{equation}\label{eqLinequlaity}
\widetilde{f}({\bolds\beta}^{t+1}) = h({\bolds\beta
}^{t+1}) + \lambda\|{\bolds\beta}^{t+1}\|_1 \leq
Q_R({\bolds\beta}^{t+1}, \bw^t).
\end{equation}
It is easy to check that when $R$ is equal to the Lipschitz constant $L$,
it will satisfy the above inequality for any ${\bolds\beta
}^{t+1}$ and
$\bw^t$. However, when it is difficult to compute the Lipschitz
constant, instead of using a global constant $L$, we could find a
sequence $\{L_t\}_{t=0}^T$ such that $L_{t+1}$ satisfies the
inequality (\ref{eqLinequlaity}) for the $t$th iteration. In
particular, we start with any small constant $L_0$. For each
iteration, we find the smallest integer $a \in\{0,1,2, \ldots\}$
such that by setting $L_{t+1}=\tau^a L_t$, where $\tau>1$ is a
predefined scaling factor, we have
%
\begin{equation}
\widetilde{f}({\bolds\beta}_{L_{t+1}}(\bw^t)) \leq
Q_{L_{t+1}}({\bolds\beta}_{L_{t+1}}(\bw^t), \bw^t).
\end{equation}
Then we set ${\bolds\beta}^{t+1}={\bolds\beta
}_{L_{t+1}}(\bw^t) \equiv\argmin Q_{L_{t+1}}({\bolds\beta},
\bw^t)$.

\subsection{Convergence rate and time complexity}
\label{subseccomplexity}

Although we optimize the approximation function $\widetilde
{f}({\bolds\beta})$ rather than the original $f({\bolds
\beta})$ directly, it can
be proven that $f(\widehat{{\bolds\beta}})$ is sufficiently
close to the optimal objective value of the \textit{original} function
$f({\bolds\beta}^*)$. The convergence rate of Algorithm \ref
{algogdglasso} is presented in the next theorem.
%
\begin{theorem}
\label{thmcomplexity} Let ${\bolds\beta}^{\ast}$ be the
optimal solution to
(\ref{eqobjglasso}) and ${\bolds\beta}^t$ be the approximate
solution at the
$t$th iteration in Algorithm \ref{algogdglasso}. If we require
$f({\bolds\beta}^t)-f({\bolds\beta}^{\ast}) \leq
\varepsilon$ where $f$ is the original objective, and set
$\mu=\frac{\varepsilon}{2D}$, then the number of iterations $t$ is
upper-bounded by
%
\begin{equation}
\label{eqbound}
\sqrt{\frac{4
\|{\bolds\beta}^{\ast}-{\bolds\beta}^0\|_2^2}{\varepsilon
}{\biggl(\lambda_{\max}(\bX^T\bX)+\frac{2D\|C\|^2}{\varepsilon
}\biggr)}}.
\end{equation}
\end{theorem}

The key\vspace*{1pt} idea behind the proof of this theorem is to decompose
$f({\bolds\beta}^t)-f({\bolds\beta}^{\ast})$ into three
parts: (i)
$f({\bolds\beta}^t)-\widetilde{f}({\bolds\beta}^{t})$, (ii)
$\widetilde{f}({\bolds\beta}^t)-\widetilde{f}({\bolds
\beta}^{\ast})$ and\vadjust{\goodbreak}  (iii)
$\widetilde{f}({\bolds\beta}^{\ast})-f({\bolds\beta
}^{\ast})$. (i) and (iii) can be bounded
by the gap of the approximation $\mu D$; and (ii) only involves
the function $\widetilde{f}$ and can be upper bounded by $O(\frac{1}{t^2})$
as shown in \citet{FISTA}. We obtain (\ref{eqbound}) by balancing
these three terms. The details of the proof are presented in the
\hyperref[app]{Appendix}. According to Theorem \ref{thmcomplexity}, Algorithm
\ref{algogdglasso} converges in $O(\frac{\sqrt{2D}}{\varepsilon})$
iterations, which is much faster than the subgradient method with
the convergence rate of $O(\frac{1}{\varepsilon^2})$. Note that the
convergence rate depends on $D$ through the term $\sqrt{2D}$, and
the $D$ depends on the problem size.
%
\begin{remark}
Since there is no line search in Algorithm \ref{algogdglasso}, we
cannot guarantee that the objective values are monotonically decreasing
over iterations theoretically. But empirically, based on our own
experience, the objective values always decrease over iterations. One
simple\vspace*{2pt} strategy to guarantee the monotone decreasing property is to
first compute $\widetilde{{\bolds\beta}}{}^{t+1}=\argmin
_{{\bolds\beta}} Q_{L}({\bolds\beta}, \bw^t)$ and then
set ${\bolds\beta}^{t+1}=\argmin_{{\bolds\beta}\in\{
\widetilde{{\bolds\beta}}{}^{t+1}, {\bolds\beta}^t\}}
f({\bolds\beta})$.
\end{remark}
%
\begin{remark}
Theorem \ref{thmcomplexity} only shows the convergence rate for the
objective value. As for the estimator ${\bolds\beta}^t$, since
it is a convex optimization problem, it is well known that
${\bolds\beta}^t$ will eventually converge to ${\bolds
\beta}^*$. However, the speed of convergence of ${\bolds\beta
}^t$ to ${\bolds\beta}^*$ depends on the structure of the input
$\bX$. If $h({\bolds\beta})$ is a~strongly convex function with
the strong convexity parameter, $\sigma>0$. In our problem, it is
equivalent to saying that $\bX^T\bX$ is a nonsingular matrix with the
smallest eigenvalue $\sigma>0$. Then we can show that if
$f({\bolds\beta}^t)-f({\bolds\beta}^*) \leq\varepsilon$ at
the convergence, then $\|{\bolds\beta}^t-{\bolds\beta}^*\|
_2 \leq\sqrt{\frac{2\varepsilon}{\sigma}}$. In other words,
${\bolds\beta}^t$ converges to~${\bolds\beta}^*$ in $\ell
_2$-distance at the rate of $O(\frac{1}{\varepsilon^2})$.
For general high-dimensional sparse learning problems with $J>N$, $\bX
^T\bX$ is singular and, hence, the optimal solution ${\bolds
\beta}^*$ is not unique. In such a case, one can only show that
${\bolds\beta}^t$ will converge to one of the optimal solutions.
But the speed of the convergence of $\|{\bolds\beta
}^t-{\bolds\beta}^*\|_2$ or its relationship with
$f({\bolds\beta}^t)-f({\bolds\beta}^*)$ is widely
recognized as an open problem in the optimization community.
\end{remark}

As for the time complexity, the main computational cost in each
iteration comes from calculating the gradient $\nabla h (\bw_t)$.
Therefore, SPG shares almost the same per-iteration time as the
subgradient descent but with a~faster convergence rate. In more
details, if $J<N$ and $\bX^T\bX$ and $\bX^T{\mathbf y}$ can be pre-computed
and stored in memory, the computation of the first part of $\nabla h
(\bw_t)$, $(\bX^T \bX) \bw_t- (\bX^T{\mathbf y})$, takes the time
complexity
of $O(J^2)$. Otherwise, if $J>N$, we can compute this part by $\bX^T
(\bX\bw_t- {\mathbf y})$, which takes the time complexity of $O(JN)$.
As for
the generic solver, IPM for SOCP for overlapping group lasso or IPM
for QP for graph-guided fused lasso, although it converges in fewer
iterations [i.e., $\log(\frac{1}{\varepsilon})$], its per-iteration
complexity is higher by orders\vadjust{\goodbreak}  of magnitude than ours as shown in
Table \ref{tabtimecomp}. In addition to time complexity, IPM
requires the pre-storage of $\bX^T\bX$ and each IPM iteration
requires significantly more memory to store the Newton linear
system. Therefore, the SPG is much more efficient and scalable for
large-scale problems.

\begin{table}
\caption{Comparison of per-iteration time complexity}
\label{tabtimecomp}
\begin{tabular*}{\tablewidth}{@{\extracolsep{\fill}}lcc@{}}
\hline
& \multicolumn{1}{c}{\textbf{Overlapping group lasso}} & \multicolumn{1}{c@{}}{\textbf{Graph-guided fused lasso}} \\
\hline
SPG & $O(J \min(J,N)+ \sum_{\giG}|g|)$ & $O(J \min(J,N)+|E|)$\\
IPM & $O((J+|\mathcal{G}|)^2(N+\sum_{\giG}|g|))$ &
$O((J+|E|)^3)$\\
\hline
\end{tabular*}
\end{table}


\subsection{Summary and discussions}

The insight of our work was drawn from two lines of earlier works. The
first one is the proximal gradient methods (e.g., Nesterov's
composite gradient method [\citet{Nesterov07}], FISTA [\citet{FISTA}].
They have been widely adopted to solve optimization problems with a
convex loss and a relatively simple nonsmooth penalty, achieving
$O(\frac{1}{\sqrt{\varepsilon}})$ convergence rate. However, the
complex structure of the nonseparable penalties considered in this
paper makes it intractable to solve the proximal operator exactly. This
is the challenge that we circumvent via smoothing.

The general idea of the smoothing technique used in this paper was
first introduced by \citet{Nesterov05}. The algorithm presented in
\citet{Nesterov05} only works for smooth problems so that it has to
smooth out the entire nonsmooth penalty. Our approach separates the
simple nonsmooth $\ell_1$-norm penalty from the complex structured
sparsity-inducing penalties. In particular, when an $\ell_1$-norm
penalty is used to enforce the \textit{individual-feature-level
sparsity} (which is especially necessary for fused lasso), we smooth
out the complex structured-sparsity-inducing penalty while leaving the
simple $\ell_1$-norm as it is. One benefit of our approach is that it
can lead to solutions with \textit{exact zeros} for irrelevant features
due to the $\ell_1$-norm penalty and hence avoid the post-processing
(i.e., truncation) step.\footnote{When there is no $\ell_1$-norm
penalty in the model (i.e., $\lambda=0$), our method still applies.
However, to conduct variable selection, as for other optimization
methods (e.g., IPM), we need a post-processing step to truncate
parameters below a certain threshold to zeros.} Moreover, the
algorithm in \citet{Nesterov05} requires the condition that
${\bolds\beta}$ is bounded and that the number of iterations is
predefined, which are impractical for real applications.\looseness=-1 

As for the convergence rate, 
the gap between $O(\frac{1}{\varepsilon})$ and the optimal
rate $O(\frac{1}{\sqrt{\varepsilon}})$ is due to the approximation of
the structured sparsity-inducing penalty. It is possible to show
that if $\bX$ has a full column rank, $O(\frac{1}{\sqrt{\varepsilon}})$
can be achieved by\vadjust{\goodbreak}  a variant of the excessive gap method
[\citet{ExcessiveGap}]. However, such a rate cannot be easily obtained
for sparse regression problems where $J>N$. For some special cases
as discussed in the next section, such as tree-structured or the
$\ell_1/\ell_{\infty}$ mixed-norm based overlapping groups,
$O(\frac{1}{\sqrt{\varepsilon}})$ can be achieved at the expense of
more computation time for solving the proximal operator. It remains
an open question whether we can further boost the
generally-applicable SPG method to achieve
$O(\frac{1}{\sqrt{\varepsilon}})$.

\section{Related optimization methods}
\label{secsummary}

\subsection{Connections with majorization--minimization}

The idea of constructing a smoothing approximation has also been
adopted in another widely used optimization method,
majorization--minimization (MM) for minimization problem (or
minorization--maximization for maximization problem) [\citet{MM}]. To
minimize a given objective,
MM replaces the difficult-to-optimize objective function with a simple
(and smooth in most cases)
surrogate function which majorizes the objective. It minimizes the
surrogate function and iterates such a procedure. The difference
between our approach and MM is that our approximation is a uniformly
smooth lower bound of the objective with a bounded gap, whereas the
surrogate function in MM is an upper bound of the objective. %
In addition, MM is an iterative procedure which iteratively constructs
and minimizes the surrogate function, while our approach constructs
the smooth approximation once and then applies the proximal gradient
descent to optimize it. With the quadratic surrogate functions for
the $\ell_2$-norm and fused-lasso penalty derived in \citet{MMLasso} and
\citet{MMFused}, one can easily apply MM to solve the structured
sparse regression problems. However, in our problems, the Hessian
matrix in the quadratic surrogate will no longer have a simple
structure (e.g., tridiagonal symmetric structure in chain-structured
fused signal approximator). Therefore, one may need to apply the
general optimization methods, for example, conjugate-gradient or
quasi-Newton method, to solve a series of quadratic surrogate
functions. In addition, since the objective functions considered in
our paper are neither smooth nor strictly convex, the local and
global convergence results for MM in \citet{MM} cannot be applied. It
seems to us still an open problem to derive the local, global
convergence and the convergence rate for MM for the general
nonsmooth convex optimization.

Recently, many first-order approaches have been developed for
various subclasses of overlapping group lasso and graph-guided fused
lasso. Below, we provide a survey of these methods:

\subsection{Related work for mixed-norm based group-lasso penalty}
\label{secrelatedgroup}

Most of the existing optimization methods developed for mixed-norm
penalties can handle only a specific subclass of the general
overlapping-group-lasso penalties. Most of these methods use the
proximal gradient framework [\citet{FISTA}, \citet{Nesterov07}]\vadjust{\goodbreak}  and focus on
the issue of how to \textit{exactly} solve the proximal operator. For
nonoverlapping groups with the $\ell_1/\ell_2$ or
$\ell_1/\ell_{\infty}$ mixed-norms, the proximal operator can be
solved via a simple projection
[\citet{Duchi09}, \citet{Liu09multi-taskfeature}]. A one-pass coordinate ascent method
has been developed
for tree-structured groups with the $\ell_1/\ell_2$ or
$\ell_1/\ell_{\infty}$ [\citet{Liu2010}, \citet{Bach10}], and quadratic
min-cost network flow for arbitrary overlapping groups with the
$\ell_1/\ell_{\infty}$ [\citet{SSparse10}].

%
\begin{sidewaystable}
\textwidth=\textheight
\tablewidth=\textwidth
\caption{Comparisons of different first-order methods for
optimizing mixed-norm based overlapping-group-lasso penalties}
\label{tabgcomp}
\begin{tabular*}{\tablewidth}{@{\extracolsep{\fill}}l c
c cccc@{}}
\hline
& \textbf{No overlap} & \textbf{No overlap} & \textbf{Overlap}
& \textbf{Overlap} & \textbf{Overlap}
& \textbf{Overlap} \\
\textbf{Method} & $\bolds{\ell_1/\ell_2}$ & $\bolds{\ell_1/\ell_{\infty}}$
& \textbf{tree} $\bolds{\ell_1/\ell
_2}$ & \textbf{tree} $\bolds{\ell_1/\ell_{\infty}}$
& \textbf{arbitrary} $\bolds{\ell_1/\ell_2}$ &
\textbf{arbitrary} $\bolds{\ell_1/\ell_{\infty}}$ \\
\hline
Projection  & $O(\frac{1}{\sqrt
{\varepsilon}})$, $O(J)$ & $O(\frac{1}{\sqrt{\varepsilon}})$, $O(J \log
J)$ & N.A. & N.A. & N.A. & N.A. \\
\quad[\citet{Liu09multi-taskfeature}]\\
[4pt]
Coordinate ascent & $O(\frac{1}{\sqrt
{\varepsilon}})$,
$O(J)$ & $O(\frac{1}{\sqrt{\varepsilon}})$, $O(J \log J)$ &
$O(\frac{1}{\sqrt{\varepsilon}})$,
& $O(\frac{1}{\sqrt{\varepsilon}})$,
 & N.A. & N.A. \\
\quad[\citet{Bach10}, & & & $O({\sum_{g \in\mathcal{G}}} |g|
)$ & $O({\sum_{g \in\mathcal{G}}} {|g|\log}|g|)$\\
\quad\citet{Liu2010}]\\
[4pt]
Network Flow [Mairal & N.A. & $O(\frac{1}{\sqrt{\varepsilon
}})$, quadratic & N.A. & $O(\frac{1}{\sqrt{\varepsilon
}})$, quadratic & N.A. & $O(\frac{1}{\sqrt{\varepsilon
}})$, quadratic \\
\quad et~al. (\citeyear{SSparse10})] & & min-cost flow & & min-cost flow  & & min-cost flow \\
[4pt]
FOBOS [Duchi and & $O(\frac{1}{\varepsilon})$, $O(J)$ & $O(\frac
{1}{\varepsilon})$, $O(J \log J)$
& $O(\frac{1}{\varepsilon})$,  &
$O(\frac{1}{\varepsilon})$,
 & $O(\frac{1}{\varepsilon^2})$,
 & $O(\frac
{1}{\varepsilon})$, quadratic  \\
\quad Singer (\citeyear{Duchi09})] & & & $O({\sum_{g \in\mathcal{G}}} |g|)$ &
$O({\sum_{g \in\mathcal{G}}} {|g|\log}|g|)$
& $O({\sum_{g \in\mathcal{G}}} |g|)$ & min-cost flow\\
&&&&& (subgradient)\\
[4pt]
SPG & $O(\frac{1}{\varepsilon})$, $O(J)$ & $O(\frac{1}{\varepsilon})$,
$O(J \log J )$
& $O(\frac{1}{\varepsilon})$,  &
$O(\frac{1}{\varepsilon})$,
& $O(\frac{1}{\varepsilon})$,  &
$O(\frac{1}{\varepsilon})$, \\
&&& $O({\sum_{g \in\mathcal{G}}} |g|) $ & $O({\sum_{g \in\mathcal{G}}} {|g|\log}|g|)$ &
$O({\sum_{g \in\mathcal{G}}} |g|)$ & $O({\sum_{g \in\mathcal{G}}} {|g|\log}|g|)$\\
\hline
\end{tabular*}
\end{sidewaystable}

Table \ref{tabgcomp} summarizes the applicability, the convergence
rate and the per-iteration time complexity for the available
first-order methods for different subclasses of group lasso
penalties. More specifically, the methods in the first three rows adopt the
proximal gradient framework. The first column of these rows gives
the solver for the proximal operator. Each entry in Table~\ref
{tabgcomp} contains the
convergence rate and the per-iteration time complexity. For the sake
of simplicity, for all methods, we omit the time for computing
the gradient of the loss function which is required for all of the
methods [i.e., $\nabla g({\bolds\beta})$ with $O(J^2)$]. The
per-iteration
time complexity in the table may come from the computation of
the proximal operator or subgradient of the penalty. ``N.A.'' stands
for ``not applicable'' or no guarantee in the convergence. 
As we can see from Table \ref{tabgcomp}, although our method is not
the most ideal one for some of the special cases, our method along
with FOBOS 
[\citet{Duchi09}] are the
only generic first-order methods that can be applied to all
subclasses of the penalties.

As we can see from Table \ref{tabgcomp}, for arbitrary overlaps
with the $\ell_1/\ell_{\infty}$, although the method proposed in
\citet{SSparse10} achieves $O(\frac{1}{\sqrt{\varepsilon}})$
convergence rate, the per-iteration complexity can be high due to
solving a~quadratic min-cost network flow problem. From the
worst-case analysis, the per-itera\-tion time complexity for solving
the network flow problem in \citet{SSparse10} is at least
$O(|V||E|)=O((J+|\mathcal{G}|)(|\mathcal{G}|+J+\sum_{g\in
\mathcal{G}}|g|))$, which is much higher than our method with
$O(\sum_{g \in\mathcal{G}} {|g|\log}|g|)$. More importantly, for the
case of arbitrary overlaps with the $\ell_1/\ell_2$, our method
has a superior convergence rate to all the other methods.

In addition to these methods, an active-set algorithm was proposed
that can be applied to the \textit{square} of the $\ell_1/\ell_2$ mixed-norm
with overlapping groups [\citet{Bach09}]. This method formulates each
subproblem involving only the active variables either as an SOCP,
which can be computationally expensive for a large active set, or as
a jointly convex problem with auxiliary variables, which is then
solved by an alternating gradient descent. The latter approach
involves an expensive matrix inversion at each iteration and lacks
the global convergence rate. Another method [\citet{Jiepinggroup}]
was proposed for the overlapping group lasso which approximately solves
the proximal operator. However, the convergence of this type of
approach cannot be guaranteed, since the error introduced in each
proximal operator will be accumulated over iterations.\vadjust{\goodbreak}  

\subsection{Related work for fused lasso}

For the graph-guided-fused-lasso penal\-ty, when the structure is a
simple chain, the pathwise coordinate descent method
[\citet{Friedman07}] can be applied. For the general graph structure,
a first-order method that approximately solves the proximal operator
was proposed in \citet{Jieping10}. However, the convergence cannot
be guaranteed due to the errors introduced in computing the proximal
operator over iterations.

Recently, two different path algorithms have been proposed [\citet
{Ryan10}, \citet{Hua11a}] that can be used to solve the graph-guided fused lasso
as a special case. Unlike the traditional optimization methods that
solve the problem for a fixed regularization parameter, they solve the
entire path of solutions, and, thus, have great practical advantages.
In addition, for both methods, updating solutions from one hitting time
to another is computationally very cheap. More specifically, a QR
decomposition based updating scheme was proposed in \citet{Ryan10} and
the updating in \citet{Hua11a} can be done by an efficient sweep operation.

However, for high-dimensional data with $J \gg N$, the path algorithms
can have the following problems:
\begin{longlist}[(1)]
\item[(1)] For a general design matrix $\bX$ other than the identity
matrix, the method in \citet{Ryan10} needs to first compute the
pseudo-inverse of $\bX\dvtx \bX^+=(\bX^T \bX)^+ \bX^T$, which could
be computationally expensive for large $J$.

\item[(2)] The original version of the algorithms in
\citet{Ryan10} and \citet{Hua11a} requires that $\bX$ has a
full column rank. When $J>N$, although one can add an extra
$\varepsilon\|{\bolds\beta}\|_2^2$ term, this changes the original
objective value especially when $\varepsilon$ is large. For smaller
$\varepsilon$, the matrix $(\bX ^*)^T\bX^*$ with $\bX^*=
\bigl[{\bX\atop \varepsilon I}\bigr]$ is highly ill-conditioned; and hence computing its inverse as
the initialization step in \citet{Ryan10} is very difficult. There is
no known result on how to balance this trade-off.

\item[(3)] In both \citet{Ryan10} and \citet{Hua11a}, the authors extend
their algorithm to deal with the case when $\bX$ does not have a full
column rank. The extended version requires a Gramm--Schmidt process as
the initialization, which could take some extra time.
\end{longlist}

In Table \ref{tabfused} we present the comparisons for different
methods. From our analysis, the method in \citet{Hua11a} is more
efficient than the one in \citet{Ryan10} since it avoids the heavy
computation of the pseudo-inverse of~$\bX$. In practice, if $\bX$ has
a full column rank and one is interested in solutions on the entire
path, the method in \citet{Hua11a} is very efficient and faster than
our method. Instead, when $J \gg N$, the path following methods may
require a time-consuming preprocessing procedure.
%
\begin{table}
\tabcolsep=0pt
\caption{Comparisons of different methods for optimizing
graph-guided fused lasso}
\label{tabfused}
\begin{tabular*}{\tablewidth}{@{\extracolsep{\fill}}lccc@{}}
\hline
\textbf{Method} & \textbf{Preprocessing}  & \textbf{Per-iteration}
& \textbf{No. of} \\
\textbf{and condition} & \textbf{time} & \textbf{time complexity}
& \textbf{iterations}\\
\hline
[Zhou and   & $O(J^3)$ & $O( (|E|+J)^2 )$ & $O(|E|+J)$ \\
\quad Lange (\citeyear{Hua11a})]  \\
\quad ($\bX$ full column \\
\quad rank, entire path)\\
[4pt]
[Tibshirani and  & \multicolumn{1}{c}{$O(J^3 + N (|E|+J) \quad\hspace*{4pt}$} &
$O(\min((|E|+J)^2, N^2))$ & $O(|E|+J)$  \\
\quad Taylor (\citeyear{Ryan10})]  &
\multicolumn{1}{c}{$\quad\hspace*{4pt}\times\min((|E|+J), N)
)$}
& & (lower bound)\\
\quad ($\bX$ full column \\
\quad rank, entire path)\\
[4pt]
[Tibshirani and   & $O(J^3 + J^2 N + (|E|+J)^2 N )$ &
$O(N^2)$ & $O(|E|+J)$  \\
\quad Taylor (\citeyear{Ryan10})] & & & (lower bound)\\
\quad($\bX$ not full column \\
\quad rank, entire path)\\
[4pt]
SPG (single & $O(NJ^2)$ & $O(J^2+|E|)$ &
$O(\frac{1}{\varepsilon})$ \\
\quad regularization\\
\quad parameter)\\
\hline
\end{tabular*}
\end{table}

\section{Extensions to multi-task regression with structures on outputs}
\label{secmulti}

\begin{figure}[b]

\includegraphics{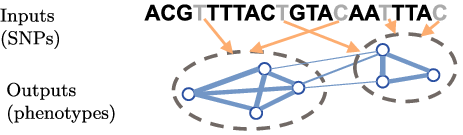}

\caption{Illustration of the multi-task regression with graph
structure on outputs.}
\label{figeQTL}
\end{figure}

The structured sparsity-inducing penalties as discussed in the
previous section can be similarly used in the multi-task
regression setting [\citet{Seyoung09}, \citet{graph09}], where the prior
structural information is available for the outputs instead of inputs.
For example,
in genetic association analysis, where the goal is to discover few
genetic variants or single nucleotide polymorphisms (SNPs) out of
millions of SNPs (inputs) that influence phenotypes (outputs) such
as gene expression measurements, the correlation structure of the
phenotypes can be naturally represented as a~graph, which can be
used to guide the selection of SNPs as shown in Figure~\ref{figeQTL}. Then, the graph-guided-fused-lasso penalty can be used to
identify SNPs that are relevant jointly to multiple related
phenotypes.

In a sparse multi-task regression with structure on the output side, we
encounter the same
difficulties of optimizing with nonsmooth and nonseparable
penalties as in the previous section, and the SPG can be extended to
this problem in a straightforward
manner. Due to the importance of this class of problems and its
applications, in this section, we
briefly discuss how our method can be applied to the multi-task
regression with structured-sparsity-inducing penalties.

\subsection{Multi-task linear regression regularized by structured
sparsity-inducing penalties}
For the simplicity of illustration, we assume all different tasks share
the same input matrix.
Let $\bX\in\mathbb{R}^{N \times J}$ denote the matrix of input
data for $J$ inputs and $\bY\in\mathbb{R}^{N \times K}$ denote the
matrix of output data for $K$ outputs over $N$ samples. We assume a
linear regression model for each of the $k$th outputs: ${\mathbf
y}_k=\bX{\bolds\beta}_k
+ \be_k,  \forall k =1, \ldots, K$, where ${\bolds\beta
}_k=[\beta_{1k},
\ldots, \beta_{Jk}]^T$ is the regression coefficient vector for the
$k$th output and $\be_k$ is Gaussian noise. Let $\bB=[{\bolds
\beta}_1, \ldots,
{\bolds\beta}_K] \in\mathbb{R}^{J \times K}$ be the matrix of
regression
coefficients for all of the $K$ outputs. Then, the multi-task (or
multivariate-response)
structured sparse regression problem can be naturally formulated as
the following optimization problem:
%
\begin{equation}\label{eqobjglassot}
\min_{\bB\in\mathbb{R}^{J \times K}} f(\bB) \equiv\frac{1}{2} \|
\bY-\bX\bB\|_F^2
+ \Omega(\bB)+\lambda\|\bB\|_1,
\end{equation}
where \mbox{$\|\cdot\|_F$} denotes the matrix Frobenius norm, \mbox{$\|\cdot\|_1$}
denotes the matrix entry-wise $\ell_1$ norm, and $\Omega(\bB)$ is a
structured sparsity-inducing penalty with a structure over the outputs.
\begin{longlist}[(1)]
\item[(1)] \textit{Overlapping-group-lasso penalty in multi-task regression.}
We define the overlapping-group-lasso penalty for a structured multi-task
regression as follows:
%
\begin{equation} \label{eqpenmultigroup}
\Omega(\bB) \equiv\gamma\sum_{j=1}^J \sum_{g\in
\mathcal{G}}w_g \|{\bolds\beta}_{jg}\|_2,
\end{equation}
where $\mathcal{G}=\{g_1,\ldots,g_{|\mathcal{G}|}\}$ is a subset of
the power set of $\{1,\ldots, K\}$ and ${\bolds\beta}_{jg}$ is
the vector of
regression coefficients corresponding to outputs in group $g\dvtx
\{\beta_{jk}, k \in g, g \in\mathcal{G}\}$. Both the
$\ell_{1}/\ell_{2}$ mixed-norm penalty for multi-task regression in
\citet{Obozinski09} and the tree-structured overlapping-group-lasso
penalty in \citet{Seyoung09} are special cases of~(\ref{eqpenmultigroup}).

\item[(2)] \textit{Graph-guided-fused-lasso penalty in multi-task regression.}
%
Assuming that a graph structure over the $K$ outputs is given as $G$
with a set of nodes $V=\{1,\ldots,K\}$, each corresponding to an
output variable and a set of edges $E$, the graph-guided-fused-lasso
penalty for a structured multi-task regression is given as
%
\begin{equation}
\Omega(\bB)=\gamma\sum_{e=(m,l)\in
E}\tau(r_{ml})\sum_{j=1}^J|{\bolds\beta}_{jm}-
\operatorname{sign}(r_{ml}){\bolds\beta}_{jl}|.
\end{equation}
\end{longlist}

\subsection{Smoothing proximal gradient descent}

%

Using similar techniques in Section \ref{secreform},
$\Omega(\bB)$ can be reformulated as
%
\begin{equation}\label{eqsaddlepent}
\Omega(\bB)= \max_{\bA\in\mathcal{Q}} \langle C\bB^T, \bA
\rangle,
\end{equation}
where $\langle\mathbf{U}, \mathbf{V} \rangle\equiv
\operatorname{Tr}(\mathbf{U}^T\mathbf{V})$ denotes a matrix inner product.
$C$ is constructed in a similar way as in (\ref{eqmatrixC}) or
(\ref{eqmatrixH}), just by replacing the index of the input
variables with the output variables, and $\bA$ is the matrix of the
auxiliary variables. 

Then we introduce the smooth approximation of
(\ref{eqsaddlepent}):
%
\begin{equation}
\label{eqfmut}
f_\mu(\bB)=\max_{\bA\in\mathcal{Q}} (\langle C\bB^T, \bA
\rangle- \mu \,d (\bA)),
\end{equation}
where $d(\bA) \equiv\frac{1}{2}\|\bA\|_F^2$. Following a proof
strategy similar to that in Theorem~\ref{thmkey}, we can show that
$f_\mu(\bB)$ is convex and smooth with gradient $\nabla f_\mu(\bB)=
(\bA^{\ast})^TC$, where $\bA^{\ast}$ is the optimal solution to
(\ref{eqfmut}). The closed-form solution of $\bA^{\ast}$ and the
Lipschitz constant for $\nabla f_\mu(\bB)$ can be derived in the
same way.

\begin{table}
\caption{Comparison of per-iteration time complexity for multi-task regression}
\label{tabtimecompmulti-task}
\begin{tabular*}{\tablewidth}{@{\extracolsep{\fill}}lcc@{}}
\hline
& \multicolumn{1}{c}{\textbf{Overlapping group lasso}} & \multicolumn{1}{c@{}}{\textbf{Graph-guided fused lasso}} \\
\hline
SPG &$O(JK\min(J,N)+ J\sum_{\giG}|g|)$ & $O(JK\min(J,N)+ J|E|)$
\\
IPM &
$O(J^2(K+|\mathcal{G}|)^2(KN+J(\sum_{g\in\mathcal
{G}}|g|)))$
& $O(J^3(K+|E|)^3)$\\
\hline
\end{tabular*}
\end{table}

By substituting $\Omega(\bB)$ in (\ref{eqobjglassot}) with
$f_\mu(\bB)$, we can adopt Algorithm \ref{algogdglasso} to solve
(\ref{eqobjglassot}) with convergence rate of
$O(\frac{1}{\varepsilon})$. The per-iteration time complexity of SPG as
compared to IPM for SOCP or QP formulation is presented in Table
\ref{tabtimecompmulti-task}. As we can see, the per-iteration
complexity for SPG is linear in $\max(|K|, \sum_{g\in\mathcal{G}}
|g|)$ or $\max(|K|, |E|)$, while traditional approaches based on IPM
scape at least cubically to the size of outputs $K$.


\section{Experiment}
\label{secexp}

In this section we evaluate the scalability and efficiency of the
smoothing proximal gradient method (SPG) on a number of structured
sparse regression problems via simulation, and apply SPG to an
overlapping group lasso problem on real genetic data.

On an overlapping group lasso problem, we compare the SPG with FOBOS
[\citet{Duchi09}] and IPM for SOCP.\footnote{We use the
state-of-the-art MATLAB package SDPT3 [\citet{SDPT3}] for SOCP.} On a
multi-task graph-guided fused lasso problem, we compare the running
time of SPG with that of the FOBOS [\citet{Duchi09}] and IPM for
QP.\footnote{We use the commercial package MOSEK
(\url{http://www.mosek.com/}) for QP. The graph-guided fused lasso can
also be solved by SOCP, but it is less efficient than QP.} Note that
for FOBOS, since the proximal operator associated with $\Omega
({\bolds\beta})$
cannot be solved exactly, we set the ``loss function'' to
$l({\bolds\beta})=g({\bolds\beta})+\Omega({\bolds
\beta})$ and the penalty to $\lambda\|{\bolds\beta}\|_1$.
According to \citet{Duchi09}, for the nonsmooth loss $l({\bolds
\beta})$,
FOBOS achieves $O(\frac{1}{\varepsilon^2})$ convergence
rate, which is slower than our method.

All experiments are performed on a standard PC with 4GB RAM and the
software is written in MATLAB. The main difficulty in comparisons is
a fair stopping criterion. Unlike IPM, SPG and FOBOS do not
generate a dual solution and, therefore, it is not possible to
compute a primal-dual gap, which is the traditional stopping
criterion for IPM. Here, we adopt a~widely used approach for
comparing different methods in the optimization literature. Since it is
well known that IPM usually gives a more accurate (i.e., lower)
objective, we set the objective obtained from IPM as the optimal
objective value and stop the first-order methods when the objective
is below 1.001 times the optimal objective. For large data sets
for which IPM cannot be applied, we stop the first-order methods
when the relative change in the objective is below $10^{-6}$. In
addition, maximum iterations are set to 20,000.


Since our main focus is on the optimization algorithm, for the purpose
of simplicity, we assume that each group in the overlapping group lasso
problem receives the same amount of regularization and, hence, set the
weights $w_g$ for all groups to be 1. In principle, more sophisticated
prior knowledge of the importance for each group can be naturally
incorporated into $w_g$. In addition, we notice that each variable $j$
with the regularization $\lambda|\beta_j|$ in $\lambda\|{\bolds
\beta}\|_1$ can be viewed as a singleton group. To ease the tuning of
parameters, we again assume that each group (including the singleton
group) receives the same amount of regularization and, hence, constrain
the regularization parameters $\lambda=\gamma$.

The smoothing parameter $\mu$ is set to
$\frac{\varepsilon}{2D}$ according to Theorem \ref{thmcomplexity},
where $D$ is determined by the problem size. It is natural that for
large-scale problems with large $D$, a larger $\varepsilon$ can be
adopted without affecting the recovery quality significantly.
Therefore, instead of setting $\varepsilon$, we directly set
$\mu=10^{-4}$, which provided us with reasonably good approximation
accuracies for different scales of problems based on our experience
for a range of $\mu$ in simulations. As for FOBOS, we set the
stepsize rate to $\frac{c}{\sqrt{t}}$ as suggested in
\citet{Duchi09}, where $c$ is carefully tuned to be $\frac{0.1}{\sqrt
{NJ}}$ for
univariate regression and $\frac{0.1}{\sqrt{NJK}}$ for multi-task
regression.


\subsection{Simulation study \textup{I}: Overlapping group lasso}

We simulate data for a univariate linear regression model with the
overlapping group structure on the\vadjust{\goodbreak}  inputs as described below. Assuming
that the
inputs are ordered, we define a sequence of groups of 100 adjacent
inputs with an overlap of 10 variables between two successive groups
so that
\[
\mathcal{G}=\bigl\{ \{1,\ldots,100\}, \{91, \ldots, 190\},
\ldots, \{J-99,\ldots, J\}\bigr\}
\]
with $J=90|\mathcal{G}|+10$. We set
$\beta_j=(-1)^j \exp(-(j-1)/100)$ for $1 \leq j \leq J$. We sample
each element of $\bX$ from i.i.d. Gaussian distribution, and
generate the output data from ${\mathbf y}=\bX{\bolds\beta}+\be
$, where $\be\sim
N(0,I_{N\times N})$.

%
\begin{table}
\caption{Comparisons of different optimization methods on the
overlapping group lasso}
\label{taboverglasso}
\begin{tabular*}{\tablewidth}{@{\extracolsep{\fill}}lcd{4.2}d{4.3}
d{4.2}d{4.3}d{4.2}d{4.3}@{}}
\hline
& &
\multicolumn{2}{c}{\textbf{$\bolds{N=1}$,000}}
& \multicolumn{2}{c}{\textbf{$\bolds{N=5}$,000}}
& \multicolumn{2}{c@{}}{\textbf{$\bolds{N=10}$,000}} \\[-4pt]
& &
\multicolumn{2}{c}{\hspace*{-2.5pt}\hrulefill}
& \multicolumn{2}{c}{\hspace*{-2.5pt}\hrulefill}
& \multicolumn{2}{c@{}}{\hspace*{-2.5pt}\hrulefill} \\
& & \multicolumn{1}{c}{\hspace*{-2.5pt}\textbf{CPU (s)}}
& \multicolumn{1}{c}{\textbf{Obj.}}
& \multicolumn{1}{c}{\hspace*{-2.5pt}\textbf{CPU (s)}} & \multicolumn{1}{c}{\textbf{Obj.}} &
\multicolumn{1}{c}{\hspace*{-2.5pt}\textbf{CPU (s)}} & \multicolumn{1}{c@{}}{\hspace*{-1pt}\textbf{Obj.}} \\
\hline
\multicolumn{8}{@{}c@{}}{$|\mathcal{G}|=10$ ($J=910$)}\\
[4pt]
$\gamma=2$ & SOCP & 103.71 & 266.683 & 493.08 &
917.132 & 3\mbox{,}777.46 & 1\mbox{,}765.518 \\
& FOBOS & 27.12 & 266.948 & 1.71 & 918.019 & 1.48 & 1\mbox{,}765.613 \\
& SPG & 0.87 & 266.947 & 0.71 & 917.463 & 1.28 & 1\mbox{,}765.692 \\
[4pt]
$\gamma=0.5$ & SOCP & 106.02 & 83.304 & 510.56 &
745.102 & 3\mbox{,}585.77 & 1\mbox{,}596.418 \\
& FOBOS & 32.44 & 82.992 & 4.98 & 745.788 & 4.65 & 1\mbox{,}597.531 \\
& SPG & 0.42 & 83.386 & 0.41 & 745.104 & 0.69 & 1\mbox{,}596.452 \\
[4pt]
\multicolumn{8}{@{}c@{}}{$|\mathcal{G}|=50$ ($J=4\mbox{,}510$)}\\
[4pt]
$\gamma=10$ & SOCP & 4\mbox{,}144.20 & 1\mbox{,}089.014 & \multicolumn{1}{c}{--} & \multicolumn{1}{c}{--}
& \multicolumn{1}{c}{--}
& \multicolumn{1}{c@{}}{--}
\\
& FOBOS & 476.91 & 1\mbox{,}191.047 & 394.75 & 1\mbox{,}533.314 & 79.82 & 2\mbox{,}263.494 \\
& SPG & 56.35 & 1\mbox{,}089.052 & 77.61 & 1\mbox{,}533.318 & 78.90 & 2\mbox{,}263.601 \\
[4pt]
$\gamma=2.5$ & SOCP & 3\mbox{,}746.43 & 277.911 & \multicolumn{1}{c}{--}
& \multicolumn{1}{c}{--} & \multicolumn{1}{c}{--} & \multicolumn{1}{c}{--}
\\
& FOBOS & 478.62 & 286.327 & 867.94 & 559.251 & 183.72 & 1\mbox{,}266.728 \\
& SPG & 33.09 & 277.942 & 30.13 & 504.337 & 26.74 & 1\mbox{,}266.723 \\
[4pt]
\multicolumn{8}{@{}c@{}}{$|\mathcal{G}|=100$ ($J=9\mbox{,}010$)}\\
[4pt]
$\gamma=20$ & FOBOS & 1\mbox{,}336.72 & 2\mbox{,}090.808 & 2\mbox{,}261.36 &
3\mbox{,}132.132 & 1\mbox{,}091.20 & 3\mbox{,}278.204 \\
& SPG & 234.71 & 2\mbox{,}090.792 & 225.28 & 2\mbox{,}692.981 & 368.52 & 3\mbox{,}278.219 \\
[4pt]
$\gamma=5$ & FOBOS & 1\mbox{,}689.69 & 564.209 & 2\mbox{,}287.11 & 1\mbox{,}302.552 & 3\mbox{,}342.61 & 1\mbox{,}185.661 \\
& SPG & 169.61 & 541.611 & 192.92 & 736.559 & 176.72 & 1\mbox{,}114.933 \\
\hline
\end{tabular*}
\end{table}

To demonstrate the efficiency and scalability of SPG, we vary $J$, $N$
and $\gamma$ and report the total CPU time in seconds and the
objective value in Table \ref{taboverglasso}. The regularization
parameter $\gamma$ is set to either $|\mathcal{G}|/5$ or $|\mathcal
{G}|/20$. As we can see from Table
\ref{taboverglasso}, first, both SPG and FOBOS are more efficient and
scalable by orders of magnitude than IPM for SOCP. For larger $J$ and
$N$, we are unable to collect the results for SOCP. 
Second, SPG is more efficient than FOBOS for almost all different
scales of the problems.\footnote{In some entries in Table \ref
{taboverglasso}, the Obj. from FOBOS is much larger than other methods.
This is because that FOBOS has reached the maximum number of iterations
before convergence. Instead, for our simulations, SPG generally
converges in hundreds of, or, at most, a few thousand, iterations and
never pre-terminates.} Third, for SPG, a smaller $\gamma$ leads
to\vadjust{\goodbreak}
faster convergence. This result is consistent with Theorem \ref
{thmcomplexity}, which shows that the number of iterations is linear in
$\gamma$ through the term $\|C\|$. Moreover, we notice that a larger
$N$ does not increase the computational time
for SPG. This is also consistent with the time complexity analysis,
which shows that for linear regression, the per-iteration time
complexity is independent of $N$.

However, we find that the solutions from IPM are more accurate and, in
fact, it is hard for first-order approaches to achieve the same
precision as IPM. Assuming that we require $\varepsilon=10^{-6}$ for
the accuracy of the solution, it takes IPM about $O(\log(\frac
{1}{\varepsilon})) \approx14$ iterations to converge, while it takes
$O(\frac{1}{\varepsilon})=10^6$ iterations for SPG. This is the drawback
for any first-order method. However, in many real applications, we do
not require the objective to be extremely accurate (e.g., $\varepsilon
=10^{-3}$ is sufficiently accurate in general) and first-order methods
are more suitable. More importantly, first-order methods can be applied
to large-scale high-dimensional problems while IPM can only be applied
to small or moderate scale problems due to the expensive computation
necessary for solving the Newton linear system.

\subsection{Simulation study \textup{II}: Multi-task graph-guided fused
lasso}

\begin{figure}

\includegraphics{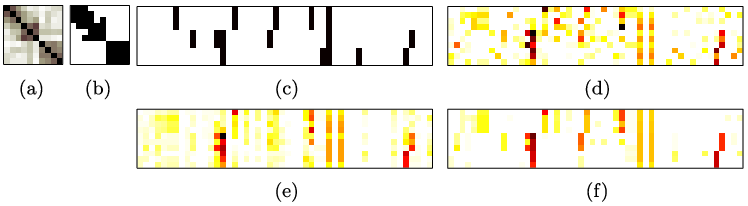}\vspace*{-3pt}

\caption{Regression coefficients estimated by
different methods based on single simulated data. $b= 0.8$ and
threshold $\rho=0.3$ for the output correlation graph are used. Red
pixels indicate large values. \textup{(a)} The correlation coefficient
matrix of phenotypes, \textup{(b)} the edges of the phenotype
correlation graph obtained at threshold 0.3 are shown as black pixels
and \textup{(c)} the true regression coefficients used in simulation.
Absolute values of the estimated regression coefficients are shown for
\textup{(d)} lasso, \textup{(e)} $\ell_1/\ell_2$ regularized multi-task
regression and \textup{(f)} graph-guided fused lasso. Rows correspond to
outputs and columns to inputs.} \label{figsimbimg}\vspace*{-3pt}
\end{figure}

We simulate data using the following scenario analogous to the
problem of genetic association mapping, where we are interested in
identifying a small number of genetic variations (inputs) that
influence the phenotypes (outputs). We use $K = 10$, $J = 30$ and $N
= 100$. To simulate the input data, we use the genotypes of the 60
individuals from the parents of the HapMap CEU panel
[\citet{hapmap2005}], and generate genotypes for an additional 40
individuals by randomly mating the original 60 individuals. We
generate the regression coefficients ${\bolds\beta}_k$'s such
that the outputs
$\mathbf{y}_k$'s are correlated with a block-like structure in the
correlation matrix. We first choose input-output pairs with nonzero
regression coefficients as we describe below. We assume three groups
of correlated output variables of sizes 3, 3 and 4. We randomly
select inputs that are relevant jointly among the outputs within
each group, and select additional inputs relevant across multiple
groups to model the situation of a higher-level correlation
structure across two subgraphs as in Figure \ref{figsimbimg}(a).
Given the sparsity pattern of $\bB$, we set all nonzero
%
$\beta_{ij}$ to a constant $b=0.8$ to construct the true coefficient
matrix $\bB$. Then, we simulate output data based on the linear
regression model with noise distributed as standard Gaussian, using
the simulated genotypes as inputs. We threshold the output
correlation matrix in Figure \ref{figsimbimg}(a) at $\rho=0.3$ to
obtain the graph in Figure \ref{figsimbimg}(b), and use this
graph as prior structural information for the graph-guided fused lasso.
As an illustrative  example, the estimated regression coefficients
from different regression models for recovering the association\vadjust{\goodbreak}
patterns are shown in Figures~\ref{figsimbimg}\mbox{(d)--(f)}. While the results of the lasso and
$\ell_1/\ell_2$-regularized multi-task regression with $\Omega(\bB)=\sum
_{j=1}^J \|{\bolds\beta}_{j,:}\|_2$ [\citet{Obozinski09}]
in Figures
\ref{figsimbimg}(d) and (e) contain many false positives, the
results from the graph-guided fused lasso in Figure
\ref{figsimbimg}(f) show fewer false positives and reveal clear
block structures. Thus, the graph-guided fused lasso proves to be a
superior regression model for recovering the true regression pattern
that involves structured sparsity in the input/output relationships.

\begin{figure}
\begin{tabular}{@{}cc@{}}

\includegraphics{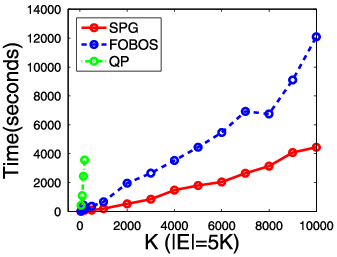}
 & \includegraphics{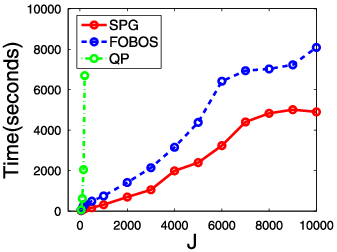}\\
(a) & (b)\\[4pt]
\multicolumn{2}{@{}c@{}}{
\includegraphics{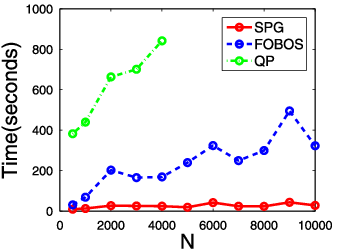}
} \\
\multicolumn{2}{@{}c@{}}{(c)}
\end{tabular}
\caption{Comparisons of SPG, FOBOS and QP.
\textup{(a)} Vary $K$ from $50$ to $10\mbox{,}000$, fixing
$N=500$, $J=100$;
\textup{(b)} vary $J$ from $50$ to $10\mbox{,}000$, fixing
$N=1\mbox{,}000$, $K=50$;
and \textup{(c)} vary~$N$ from $500$ to $10\mbox{,}000$,
fixing $J=100$, $K=50$.}
\label{figsynscalegraph}
\end{figure}

To compare SPG with FOBOS and IPM for QP in solving such a structured
sparse regression problem, we vary $K$, $J$, $N$ and present the
computation time in seconds in Figures \ref{figsynscalegraph}(a)--(c),
respectively. We select the regularization parameter $\gamma$ using
separate validation data, and report the CPU time for the graph-guided
fused lasso with the selected $\gamma$. The input/output data and true
regression coefficient matrix $\bB$
%
are generated in a way similar as above. More precisely, we assume
that each group of correlated output variables is of size 10. For
each group of the outputs, we randomly select $10\%$ of the input
variables as relevant. In addition, we randomly select $5\%$ of the
input variables as relevant to every two consecutive groups of
outputs and $1\%$ of the input variables as relevant to every three
consecutive groups. We set the $\rho$ for each data item so that the
number of edges is 5 times the number of the nodes (i.e., $|E|=5K$).
Figure \ref{figsynscalegraph} shows that SPG is
substantially more efficient and can scale up to very
high-dimensional and large-scale data sets. Moreover, we notice that
the increase of $N$ almost does not affect the computation time of
SPG, which is consistent with the complexity analysis in
Section \ref{subseccomplexity}.

\subsection{Real data analysis: Pathway analysis of breast cancer data}

In this section we apply the SPG to an overlapping group lasso
problem with a~logistic loss on real-world data collected from\vadjust{\goodbreak}
breast cancer tumors [\citet{Jacob09}, \citet{Marcgene02}]. 
The main goal is to demonstrate the importance of employing
structured sparsity-inducing penalties for performance enhancement
in real life high-dimensional regression problems, thereby further
exhibiting and justifying the needs of efficient solvers such as SPG for
such problems.

The data are given as gene expression measurements for 8,141 genes
in 295 breast-cancer tumors (78 metastatic and 217 nonmetastatic).
A lot of research efforts in biology have been devoted to identifying
biological pathways that consist of a group of genes participating in a
particular biological
process to perform a certain functionality in the cell.
Thus, a powerful way of discovering genes involved in a tumor growth is
to consider groups of interacting genes in each pathway rather than
individual genes independently [\citet{breastpathway}].
The overlapping-group-lasso penalty provides us with a natural way to
incorporate this known pathway information into the biological analysis,
where each group consists of the genes in each pathway. This approach
can allow us to find pathway-level gene groups of
significance that can distinguish the two tumor types.
In our analysis of the breast cancer data, we cluster the genes using
the canonical pathways from the Molecular Signatures Database
[\citet{MSigDB}],
and construct the overlapping-group-lasso penalty using the
pathway-based clusters as groups.
Many of the groups overlap because genes can participate in multiple pathways.
Overall, we obtain 637 pathways over 3,510 genes, with each pathway
containing 23.47 genes on average and each gene appearing in four
pathways on average.
Instead of analyzing all 8,141 genes, we focus on these 3,510 genes
which belong to certain pathways. We set up the optimization problem of
minimizing the logistic loss with the overlapping-group-lasso penalty
to classify the tumor types based on the gene expression levels,
and solve it with SPG.

Since the number of positive and negative samples are
imbalanced, we adopt the balanced error
rate defined as the average error rate of the two classes.\footnote{See
\url{http://www.modelselect.inf.ethz.ch/evaluation.php} for more
details.}
We split the data into the training and testing sets with the ratio of
\mbox{$2:1$}, and vary the $\lambda=\gamma$ from large to small to obtain
the full regularization path.

\begin{figure}
\begin{tabular}{cc}

\includegraphics{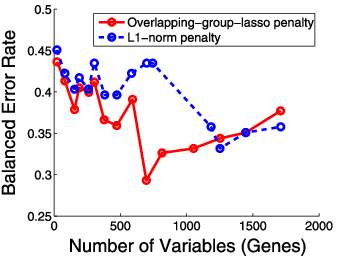}
 & \includegraphics{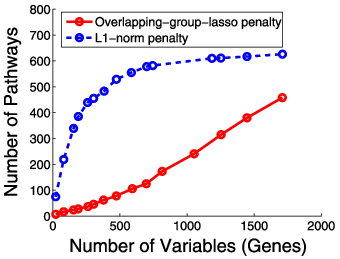}\\
(a) & (b)
\end{tabular}
\caption{Results from the analysis of breast cancer data.
\textup{(a)} Balanced error rate for varying the number of selected genes,
and \textup{(b)} the number of pathways for varying the number of selected
genes.}
\label{figcancerpathway}
\end{figure}

In Figure \ref{figcancerpathway} we compare the results from
fitting the logistic regression with the overlapping-group-lasso
penalty with a baseline model with only the $\ell_1$-norm penalty. Figure
\ref{figcancerpathway}(a) shows the balanced error rates for the
different numbers of selected genes along the regularization path.
As we can see, the balanced error rate for the model with the
overlapping-group-lasso penalty is lower than the one with the
$\ell_1$-norm, especially when the number of selected genes is
between 500 to 1,000. The model with the overlapping-group-lasso
penalty achieves the best error rate of 29.23\% when 696 genes are
selected, and these 696\vadjust{\goodbreak} genes belong to 125 different pathways. In
Figure \ref{figcancerpathway}(b), for the different numbers of
selected genes, we show the number of pathways to which the selected
genes belong. From Figure~\ref{figcancerpathway}(b) we see that
when the group structure information is incorporated, fewer pathways
are selected. This indicates that regression with the
overlapping-group-lasso penalty selects the genes at the pathway
level as a functionally coherent group, leading to an easy
interpretation for functional analysis. On the other hand, the genes
selected via the $\ell_1$-norm penalty are scattered across many
pathways, as genes are considered independently for selection. The
total computational time for computing the whole regularization path
with 20 different values for the regularization parameters is 331
seconds for the overlapping group lasso.\looseness=-1

We perform functional enrichment analysis on the selected pathways,
using the functional annotation tool [\citet{Davidtool}], and verify that
the selected pathways are significant in their relevance to the
breast-cancer tumor types.
For example, in a highly sparse model obtained with the group-lasso
penalty at the very left end of Figure \ref{figcancerpathway}(b),
the selected gene markers belong to only seven pathways, and many
of these pathways appear to be reasonable candidates for an
involvement in breast cancer.
For instance, all proteins in one of the selected
pathways are involved in the activity of \textit{proteases},
whose function is to degrade unnecessary or damaged proteins
through a chemical reaction that
breaks peptide bonds. 
One of the most important malignant properties of cancer involves
the uncontrolled growth of a group of cells, and \textit{protease
inhibitors}, which degrade misfolded proteins, have been extensively
studied in the treatment of cancer. Another interesting pathway
selected by the overlapping group lasso is known for its involvement in
\textit{nicotinate and nicotinamide metabolism}. This pathway has been
confirmed as a marker for breast cancer in previous studies
[\citet{breastpathway}]. In particular, the gene \textit{ENPP1}
(ectonucleotide pyrophosphatase/phosphodiesterase 1) in this pathway
has been found to be overly expressed in breast tumors
[\citet{ENPP1}].
Other selected pathways include the one related to ribosomes and
another related
to DNA polymerase, which are critical in the process of generating
proteins from DNA and relevant to the property of
uncontrolled growth in cancer cells.

We also examine the number of selected pathways that give
the lowest error rate in Figure \ref{figcancerpathway}.
At the error rate of 29.23\%, 125 pathways (696 genes) are
selected. It is interesting to notice that among these 125 pathways,
one is closely related to \textit{apoptosis}, which is the process of
programmed cell death that occurs in multicellular organisms
and is widely known to be involved in uncontrolled tumor growth in cancer.
Another pathway involves the genes \textit{BRCA1}, \textit{BRCA2} and
\textit{ATR}, which have all been associated with cancer susceptibility.

For comparison, we examine the genes selected with the $\ell_1$-norm
penalty that does not consider the pathway information. In this
case, we do not find any meaningful functional enrichment signals
that are relevant to breast cancer.\vadjust{\goodbreak} For example, among the 582
pathways that involve 687 genes at 37.55\% error rate, we find two
large pathways with functional enrichments, namely, \textit{response to
organic substance} (83 genes with $p$-value 3.3E$-$13) and the
process of \textit{oxidation reduction} (73~genes with $p$-value
1.7E$-$11). However, both are quite large groups and matched to
relatively high-level biological processes that do not provide much
insight on cancer-specific pathways.\vspace*{-3pt}

\section{Conclusions and future work}
\label{secconclusion}

In this paper we investigated an optimization problem for estimating
the structured-sparsity pattern in regression coefficients under a
general class of structured sparsity-inducing penalties. Many of the
structured sparsity-inducing penalties including the
overlapping-group-lasso penalties and graph-guided-fused-lasso penalty
share a common set of difficulties in optimization such as
nonseparability and nonsmoothness. We showed that the optimization
problems with these penalties can be transformed into a common
form, and proposed a general optimization approach, called the smoothing
proximal gradient method, for efficiently solving the optimization
problem of this common form. Our results show that the proposed
method enjoys both desirable theoretical guarantee and practical
scalability under various difficult settings involving complex
structure constraints, multi-task and high-dimensionality.

There are several future directions for this work. First, it is known
that reducing $\mu$ over iterations leads to better empirical results.
However, in such a scenario, the convergence rate is harder to analyze.
Moreover, since the method is only based on gradient, its online
version with the stochastic gradient descent can be easily derived.
However, proving the regret bound will require a more careful investigation.

Another interesting direction is to incorporate other accelerating
techniques into our method to further boost the performance. For
example, the technique introduced in \citet{Hua11b} can efficiently
accelerate the algorithms which essentially solve a fixed point
problem as ${\bolds\beta}=F({\bolds\beta})$. It uses an
approximation of the Jacobian of
$F({\bolds\beta})$. It is very interesting to incorporate this
technique into
our framework. However, since there is an $\ell_1$-norm penalty in
our model and the operator $F$ is hence nondifferentiable, it is
difficult to compute the approximation of the Jacobian of~$F$. One
potential strategy is to use the idea from the semi-smooth Newton method
[\citet{Newton93}, \citet{Defeng02}] to solve the nonsmooth operator $F$.\vspace*{-3pt}

\begin{appendix}\label{app}
\section*{Appendix}

\subsection{\texorpdfstring{Proof of Theorem \protect\ref{thmkey}}{Proof of Theorem 1}}\vspace*{-3pt}

We first introduce the concept of \textit{Fenchel conjugate}.

\begin{definition}\label{defconj}
The Fenchel conjugate of a function $\varphi({\bolds\alpha})$
is the function~$\varphi^{\ast}({\bolds\beta})$ defined as
\[
\varphi^{\ast}({\bolds\beta})= \sup_{{\bolds\alpha}\in
\operatorname{dom}(\varphi)}
\bigl({\bolds\alpha}^T {\bolds\beta}- \varphi
({\bolds\alpha})\bigr).\vadjust{\goodbreak}
\]
\end{definition}

Recall\vspace*{1pt} that $d({\bolds\alpha})=\frac{1}{2}\|{\bolds\alpha
}\|^2$ with the
$\operatorname{dom}({\bolds\alpha})=\mathcal{Q}$. According to
Definition~\ref{defconj}, the conjugate of $d(\cdot)$ at $\frac{C{\bolds
\beta}}{\mu}$
is $ d^{\ast}(\frac{C{\bolds\beta}}{\mu})=\sup
_{{\bolds\alpha}\in\mathcal{Q}} ({\bolds\alpha}^T
\frac{C {\bolds\beta}}{\mu} -
d({\bolds\alpha})) $
and, hence,
\[
f_{\mu}({\bolds\beta}) \equiv\argmax_{{\bolds\alpha
}\in\mathcal{Q}} \bigl(
{\bolds\alpha}^T C {\bolds\beta}- \mu d( {\bolds
\alpha}) \bigr) = \mu
d^{\ast}\biggl(\frac{C{\bolds\beta}}{\mu} \biggr).
\]
According to Theorem 26.3
in \citet{Rock96}, ``a closed proper convex function is essentially
strictly convex if and only if its conjugate is essentially
smooth.'' Since $d({\bolds\alpha})$ is a closely proper strictly convex
function, its conjugate is smooth. Therefore, $f_{\mu}({\bolds
\beta})$ is a
smooth function.

Now we apply Danskin's theorem [Proposition B.25 in \citet{Ber99}] to
derive $\nabla f_{\mu} ({\bolds\beta})$. Let $\phi
({\bolds\alpha}, {\bolds\beta})= {\bolds\alpha}^T
C {\bolds\beta}-
\mu \,d( {\bolds\alpha})$. Since $d(\cdot)$ is a strongly convex function,
$\argmax_{{\bolds\alpha}\in\mathcal{Q}} \phi({\bolds
\alpha}, {\bolds\beta})$ has a unique optimal
solution and we denote it as~${\bolds\alpha}^{\ast}$. According
to Danskin's
theorem,
%
\begin{equation}
\nabla f_{\mu} ({\bolds\beta}) = \nabla_{{\bolds\beta}}
\phi({\bolds\alpha}^{\ast}, {\bolds\beta})= C^T
{\bolds\alpha}^{\ast}.
\end{equation}

As for the proof of the Lipschitz constant of $f_\mu({\bolds
\beta})$, readers may
refer to \citet{Nesterov05}.

\subsection{\texorpdfstring{Proof of Proposition \protect\ref{propgroup}}{Proof of Proposition 1}}

\begin{eqnarray}
\label{eqbastar} {\bolds\alpha}^{\ast} & = & \argmax
_{{\bolds\alpha}\in\mathcal{Q}}
\biggl( {\bolds\alpha}^T C {\bolds\beta}
- \frac{\mu}{2}\|{\bolds\alpha}\|_2^2\biggr) \nonumber\\
& = & \argmax_{{\bolds\alpha}\in\mathcal{Q}} \sum_{g \in
\mathcal{G}}
\biggl(\gamma w_g {\bolds\alpha}_g^T {\bolds\beta}_g -
\frac{\mu}{2} \|{\bolds\alpha}_g\|_2^2 \biggr)
\\
& = & \argmin_{{\bolds\alpha}\in\mathcal{Q}} \sum_{g \in
\mathcal{G}}
\biggl\|{\bolds\alpha}_g-\frac{\gamma w_g {\bolds\beta}_g}{\mu
}\biggr\|_2^2 \nonumber.
\end{eqnarray}
Therefore, (\ref{eqbastar}) can be decomposed into $|\mathcal{G}|$
independent problems: each one is the Euclidean projection onto the
$\ell_2$-ball:
\[
{\bolds\alpha}^{\ast}_g= \argmin_{{\bolds\alpha}_g\dvtx \|
{\bolds\alpha}_g\|_2 \leq1} \biggl\|{\bolds\alpha}_g-\frac
{\gamma w_g
{\bolds\beta}_g}{\mu}\biggr\|_2^2
\]
and ${\bolds\alpha}^{\ast}=[({\bolds\alpha
}_{g_1}^{\ast})^T, \ldots,
({\bolds\alpha}_{g_{|\mathcal{G}|}}^{\ast})^T]^T$.
According to the property of the $\ell_2$-ball, it can be easily shown
that
\[
{\bolds\alpha}^{\ast}_g= S\biggl(\frac{\gamma w_g {\bolds\beta
}_g}{\mu}\biggr),
\]
where
\[
S(\bu) =
\cases{
\dfrac{\bu}{\|\bu\|_2}, &\quad $\|\bu\|_2
> 1$,\vspace*{2pt}\cr
\bu, &\quad $\|\bu\|_2 \leq1$.}
\]

As for $\|C\|$,
%
\begin{eqnarray*}
\|C\bv\|_2 =\gamma\sqrt{\sum_{\giG} \sum_{j\in g}(w_g)^2
v_{j}^2} =\lambda\sqrt{\sum_{j=1}^J \biggl( \sum_{\giG\ \mathrm{s.t.}\  j\in
g}(w_g)^2 \biggr) v_j^2},
\end{eqnarray*}
the maximum value of $\|C\bv\|_2$, given
$\|\bv\|_2 \leq1$, can be achieved by setting~$v_{\hat{j}}$ for
$j$ corresponding to the largest summation \mbox{$\sum_{\giG\ \mathrm{s.t.}\ j\in
g}(w_g)^2$} to one, and setting other $v_j$'s to zeros. Hence, we
have
\[
\|C\bv\|_2=\gamma\max_{j \in\{1, \ldots, J\}}
\sqrt{\sum_{\giG\ \mathrm{s.t.}\ j\in g} (w_g)^2}.
\]

\subsection{\texorpdfstring{Proof of Proposition \protect\ref{propnormgraph}}{Proof of Proposition 2}}

Similar to the proof technique of Proposition \ref{propgroup}, we reformulate the
problem of solving ${\bolds\alpha}^{\ast}$ as a Euclidean projection:
\[
{\bolds\alpha}^{\ast} = \argmax_{{\bolds\alpha}\in
\mathcal{Q}}
\biggl( {\bolds\alpha}^T C {\bolds\beta}- \frac{\mu
}{2}\|{\bolds\alpha}\|_2^2\biggr)
= \argmin_{{\bolds\alpha}\dvtx\|{\bolds\alpha}\|_{\infty
}\leq1} \biggl\|{\bolds\alpha}-
\frac{C{\bolds\beta}}{\mu}\biggr\|_2^2,
\]
and the\vspace*{1pt} optimal solution ${\bolds\alpha}^{\ast}$ can be
obtained by projecting
$\frac{C{\bolds\beta}}{\mu}$ onto the $\ell_{\infty}$-ball.

According to
the construction of the matrix $C$, we have, for any vector $\bv$,
%
\begin{equation}
\label{eqapp1} \|C \bv\|_2^2=\gamma^2 \sum_{e=(m,l)\in
E}(\tau(r_{ml}))^2\bigl(v_{m}-\operatorname{sign}(r_{ml})v_{l}\bigr)^2.
\end{equation}

By the simple fact that $(a \pm b)^2\leq2a^2+2b^2$ and the
inequality holds as equality if and only if $a= \pm b$, for each
edge $e=(m,l)\in E$, the value $(v_{m}-\operatorname{sign}(r_{ml})v_{l})^2$
is upper bounded by $2v_{m}^2+2v_{l}^2$. Hence, when $\|\bv\|_2=1$,
the right-hand side of (\ref{eqapp1}) can be further bounded by
%
\begin{eqnarray}
\label{eqapp2}
\|C \bv\|_2^2& \leq &\gamma^2\sum_{e=(m,l)\in E}2(\tau
(r_{ml}))^2(v_m^2+v_l^2)\nonumber\\
&=&\gamma^2\sum_{j \in V}\biggl(\sum_{e\ \mathrm{incident on}\ k}2(\tau
(r_{e}))^2\biggr) v_j^2\nonumber\\[-8pt]\\[-8pt]
&=&\gamma^2\sum_{j \in V} 2d_j v_j^2\nonumber\\
&\leq& 2\gamma^2 \max_{j \in V}d_j,
\nonumber
\end{eqnarray}
where
\[
d_j=\sum_{e\in E \ \mathrm{s.t.} \  e \ \mathrm{incident}\ \mathrm{on} \
j}(\tau(r_e))^2.
\]
Therefore, we have
\[
\|C\|\equiv\max_{\|\bv\|_2 \leq1 } \|C\bv\|_2 \leq\sqrt{2\gamma
^2 \max_{j \in V}d_j}.
\]

Note that this upper bound is tight because the first inequality in
(\ref{eqapp2}) is tight.

\subsection{\texorpdfstring{Proof of Theorem \protect\ref{thmcomplexity}}{Proof of Theorem 2}}

Based on the result from \citet{FISTA}, we have the following lemma:
%
\begin{lemma}
\label{lemsmooth} For the function $\widetilde{f}({\bolds\beta
})=h({\bolds\beta})+\lambda
\|{\bolds\beta}\|_1$, where $h({\bolds\beta})$ is an
arbitrary convex smooth function
and its gradient $\nabla h ({\bolds\beta})$ is Lipschitz
continuous with the
Lipschitz constant $L$, we apply Algorithm \ref{algogdglasso} to
minimize $\widetilde{f}({\bolds\beta})$ and let ${\bolds
\beta}^t$ be the approximate
solution at the $t$th iteration. For any ${\bolds\beta}$, we
have the
following bound:
%
\begin{equation}
\label{eqsmoothbound}
\widetilde{f}({\bolds\beta}^t)-\widetilde{f}({\bolds
\beta}) \leq\frac{2L\|{\bolds\beta}-{\bolds\beta}^0\|
_2^2}{t^2}.
\end{equation}
\end{lemma}

In order to use the bound in (\ref{eqsmoothbound}), we use the
similar proof scheme as in \citet{Lan10} and decompose
$f({\bolds\beta}^t)-f({\bolds\beta}^{\ast})$ into three terms:
%
\begin{eqnarray}
\label{eqdecompose}
f({\bolds\beta}^t)-f({\bolds\beta}^{\ast})&=&
\bigl(f({\bolds\beta}^t)-\widetilde{f}({\bolds\beta}^t)
\bigr) + \bigl( \widetilde{f}({\bolds\beta}^t)
-\widetilde{f}({\bolds\beta}^{\ast})\bigr)\nonumber\\[-8pt]\\[-8pt]
&&{} + \bigl(
\widetilde{f}({\bolds\beta}^{\ast}) - f({\bolds\beta
}^{\ast})\bigr).\nonumber
\end{eqnarray}

According to the definition of $\widetilde{f}$, we know that for any
${\bolds\beta}$
\[
\widetilde{f}({\bolds\beta}) \leq f({\bolds\beta}) \leq
\widetilde{f}({\bolds\beta}) + \mu D,
\]
where $D \equiv\max_{{\bolds\alpha}\in\mathcal
{Q}}d({\bolds\alpha})$. Therefore, the
first term in (\ref{eqdecompose}), $f({\bolds\beta
}^t)-\widetilde{f}({\bolds\beta}^t)$, is
upper-bounded by $\mu D$, and the last term in (\ref{eqdecompose})
is less than\vspace*{1pt} or equal to~0 [i.e., $\widetilde{f}({\bolds\beta
}^{\ast}) -
f({\bolds\beta}^{\ast}) \leq0$]. Combining (\ref
{eqsmoothbound}) with these
two simple bounds, we have
%
\begin{eqnarray}
\label{eqbound1}
f({\bolds\beta}^t)-f({\bolds\beta}^{\ast}) &\leq&\mu D +
\frac{2L\|{\bolds\beta}^{\ast}-{\bolds\beta}^0\|
_2^2}{t^2} \nonumber\\[-8pt]\\[-8pt]
&\leq&\mu D +
\frac{2\|{\bolds\beta}^{\ast}-{\bolds\beta}^0\|
_2^2}{t^2} \biggl(\lambda_{\max}(\bX^T\bX) + \frac{\|C\|^2}{\mu
}\biggr).\nonumber
\end{eqnarray}
By setting $\mu=\frac{\varepsilon}{2D}$ and plugging this into the
right-hand side of (\ref{eqbound1}), we obtain
%
\begin{equation}
\label{eqbound2}
f({\bolds\beta}^t)-f({\bolds\beta}^{\ast}) \leq\frac
{\varepsilon}{2}
+\frac{2\|{\bolds\beta}^{\ast}\|_2^2}{t^2}\biggl(\lambda
_{\max}(\bX^T\bX)+\frac{2D\|C\|^2}{\varepsilon}\biggr).
\end{equation}
If we require the right-hand side of (\ref{eqbound2}) to be equal
to $\varepsilon$ and solve it for $t$, we obtain the bound of $t$ in
(\ref{eqbound}).
\end{appendix}

%
%

%
%

\section*{Acknowledgments}

We would like to thank Yanjun Qi for the help of preparation and
verification of breast cancer data, and Javier Pe\~na for the discussion
of the related first-order methods. We would also like to thank the
anonymous reviewers and the Associate Editor for their constructive
comments on improving the quality of the paper.


%

\printaddresses

\end{document}